\documentclass{article}

\PassOptionsToPackage{numbers, compress}{natbib}
\usepackage[preprint]{neurips_2025}

\usepackage[utf8]{inputenc} % allow utf-8 input
\usepackage[T1]{fontenc}    % use 8-bit T1 fonts
\usepackage{hyperref}       % hyperlinks
\usepackage{url}            % simple URL typesetting
\usepackage{booktabs}       % professional-quality tables
\usepackage{amsfonts}       % blackboard math symbols
\usepackage{nicefrac}       % compact symbols for 1/2, etc.
\usepackage{microtype}      % microtypography
\usepackage[dvipsnames]{xcolor}
\usepackage[most]{tcolorbox}
\tcbuselibrary{skins}

\usepackage{amsmath}
\usepackage{amssymb}
\usepackage{mathtools}
\usepackage{amsthm}
\usepackage{extarrows}
\usepackage{float}

\usepackage{graphicx}
\usepackage[utf8]{inputenc} % allow utf-8 input
\usepackage[T1]{fontenc}    % use 8-bit T1 fonts
\usepackage{hyperref}       % hyperlinks
\usepackage{url}            % simple URL typesetting
\usepackage{booktabs}       % professional-quality tables
\usepackage{amsfonts}       % blackboard math symbols
\usepackage{nicefrac}       % compact symbols for 1/2, etc.
\usepackage{microtype}      % microtypography
\usepackage{xcolor}         % colors
\usepackage{multirow}
\theoremstyle{plain}
\newtheorem{theorem}{Theorem}[section]

\theoremstyle{definition}
\newtheorem{definition}[theorem]{Definition}

\theoremstyle{remark}

\usepackage{xspace}
\usepackage{wrapfig}
\usepackage{dsfont}
\usepackage{colortbl}
\definecolor{verylightgray}{gray}{0.98}

\newcommand{\GCN}{\textit{GCN}\xspace}
\newcommand{\GAT}{\textit{GAT}\xspace}
\newcommand{\GAE}{\textit{GAE}\xspace}
\newcommand{\SAGE}{\textit{SAGE}\xspace}

\newcommand{\BUDDY}{\textit{BUDDY}\xspace}
\newcommand{\ELPH}{\textit{ELPH}\xspace}
\newcommand{\NEOGNN}{\textit{Neo-GNN}\xspace}
\newcommand{\NCN}{\textit{NCN}\xspace}
\newcommand{\NCNC}{\textit{NCNC}\xspace}
\newcommand{\SEAL}{\textit{SEAL}\xspace}

\newcommand{\SyntData}{\texttt{LR-EXP}\xspace}

\newcommand{\meanstd}[2]{#1 {\scriptsize ($\pm$ #2)}}

\usepackage{enumitem}

\newcommand{\first}[1]{\textcolor{RoyalBlue}{\textbf{#1}}}
\newcommand{\second}[1]{\textcolor{TealBlue}{\textbf{#1}}}
\newcommand{\third}[1]{\textcolor{Orange}{\textbf{#1}}}

\title{Bridging Theory and Practice in Link Representation with Graph Neural Networks}

\author{%
  Veronica Lachi\thanks{Equal contribution} \\
  Fondazione Bruno Kessler\\
  Trento, Italy \\
  \texttt{vlachi@fbk.eu} \\
  \And
  Francesco Ferrini\footnotemark[1] \\
  University of Trento \\
  Trento, Italy \\
  \texttt{francesco.ferrini@unitn.it} \\
  \And
  Antonio Longa \\
  University of Trento \\
  Trento, Italy \\
  \texttt{antonio.longa@unitn.it} \\
  \And
  Bruno Lepri \\
  Fondazione Bruno Kessler \\
  Trento, Italy \\
  \texttt{lepri@fbk.eu} \\
  \And
  Andrea Passerini \\
  University of Trento \\
  Trento, Italy \\
  \texttt{andrea.passerini@unitn.it} \\
  \And 
  Manfred Jaeger \\
  University of Aalborg \\
  Aalborg, Denmark \\
  \texttt{jaeger@cs.aau.dk} \\
}

\begin{document}

\maketitle

\begin{abstract}
Graph Neural Networks (GNNs) are widely used to compute representations of node pairs for downstream tasks such as link prediction. Yet, theoretical understanding of their expressive power has focused almost entirely on graph-level representations. In this work, we shift the focus to links and provide the first comprehensive study of GNN expressiveness in link representation. We introduce a unifying framework, the $k_\phi$-$k_\rho$-$m$ framework, that subsumes existing message-passing link models and enables formal expressiveness comparisons. Using this framework, we derive a hierarchy of state-of-the-art methods and offer theoretical tools to analyze future architectures. To complement our analysis, we propose a synthetic evaluation protocol comprising the first benchmark specifically designed to assess link-level expressiveness. Finally, we ask: does expressiveness matter in practice? We use a graph symmetry metric that quantifies the difficulty of distinguishing links and show that while expressive models may underperform on standard benchmarks, they significantly outperform simpler ones as symmetry increases, highlighting the need for dataset-aware model selection.
\end{abstract}

\section{Introduction}

Graph Neural Networks (GNNs) have achieved remarkable success across a wide range of tasks involving structured data, including node-level ~\citep{ferrini2024meta,xiao2022graph,maurya2022simplifying,kipf2016semi}, graph-level~\citep{errica2019fair,xie2022active}, and link-level tasks~\citep{Srinivasan2020On,zhang2021labeling,yun2021neo,chamberlaingraph}.
% In such tasks, the goal is to learn meaningful representations for links in a graph, which are then used for impactful applications, which include recommender systems~\citep{ying2018graph}, knowledge graph completion~\citep{nickel2015review}, and biological interaction prediction~\citep{jha2022prediction}. 
Despite the growing importance of link representation learning, the theoretical analysis of GNNs has so far focused almost exclusively on their expressiveness for graph-level representation~\citep{xu2018how,morris2019weisfeiler,bevilacquaequivariant}. In particular, it is well-established that message-passing GNNs are at most as powerful as the 1-dimensional Weisfeiler-Lehman (1-WL) test in distinguishing non-isomorphic graphs \citep{xu2018how, morris2019weisfeiler}. However, their ability to distinguish \emph{links} -- that is, to generate discriminative representations of node pairs -- remains far less understood. 

It is well known that standard GNNs struggle to distinguish structurally different links, as they typically compute link embeddings by aggregating the representations of the two endpoint nodes. This node-centric strategy introduces a significant expressiveness bottleneck: links whose endpoints are automorphic may be mapped to identical representations, even if their structural roles in the graph differ~\citep{Srinivasan2020On,zhang2021labeling,chamberlaingraph}. To mitigate this issue, several methods have extended GNNs with structural features (SFs)~\citep{zhang2021labeling,yun2021neo,wangneural,chamberlaingraph}. While such approaches are more expressive than standard GNNs, there is still no unifying framework to systematically characterize their discriminative power or organize them into a principled expressiveness hierarchy—unlike the case of graph-level representation, where the $k$-Weisfeiler-Lehman hierarchy provides a widely adopted standard~\citep{morris2019weisfeiler,xu2018how}.

In this work, we bridge this gap by conducting a comprehensive investigation into the expressive power of GNN-based models for \emph{link representation}. We observe that most existing methods for link representation combine two ingredients: a message-passing GNN for node encoding and a pairwise function for aggregating information about the target link. Building on this insight, we propose a \emph{general theoretical framework} that captures a broad class of link representation methods. The framework characterizes models based on two key dimensions: the expressiveness of the functions involved and the radius of the neighborhood around the link that they access. It subsumes many state-of-the-art models and allows us to formally reason about their ability to distinguish links.

We explore link-level expressiveness along three axes. \textbf{First}, grounded in the proposed framework, we develop a theoretical foundation to analyze and \emph{compare the expressiveness of link representation models}. Specifically, we derive criteria to assess a model’s capacity to discriminate between different link structures, enabling us to organize existing methods into an expressiveness hierarchy and to position novel models within it. \textbf{Second}, we introduce a \emph{synthetic evaluation protocol} designed to probe the expressiveness of link representation methods. This protocol comprises both a synthetic dataset and an accompanying evaluation procedure. While previous work has introduced synthetic tools for assessing graph-level expressiveness \citep{bianchi2023expressive,wang2023empirical,abboud2020surprising,murphy2019relational,balcilar2021breaking,bouritsas2022improving}, to the best of our knowledge, no analogous setup exists for links. Our procedure enables systematic, reproducible, and scalable comparisons across models. \textbf{Third}, we investigate the \emph{practical relevance} of expressiveness on link representation by studying whether more expressive models lead to better performance in real-world scenarios. We propose a graph symmetry metric to quantify structural ambiguity among links, and we empirically show that while simple models suffice in low-symmetry settings, more expressive models significantly outperform them as symmetry increases. 

%Our contributions are summarized as follows:
% \begin{itemize}
%     \item We propose a unifying theoretical framework that encompasses a wide range of GNN-based link representation methods and enables formal expressiveness analysis.
%     \item Based on this framework, we derive theoretical criteria that allow us to compare and hierarchically organize link representation models by their expressiveness.
%     \item We introduce a synthetic evaluation protocol -- including a dataset and an evaluation method -- specifically designed to quantify the expressiveness of link representation approaches.
%     \item We empirically demonstrate that expressiveness correlates with improved performance on real-world link prediction benchmarks characterized by high graph symmetry.
% \end{itemize}

\section{Preliminaries}

\begin{definition}[\textit{graph}]\label{graph}
A \textbf{graph} is a tuple $G=(V_G ,E_G)$ where $V_G =\{1,\ldots,n\}$ is a set of nodes, $E_G \subseteq V \times V $ is a set of edges. To each graph is associated a node features matrix  $\mathbf{X}^0\in \mathbb{R}^{n\times f}$ and an adjacency matrix $\mathbf{A}_G \in \{0,1\}^{n\times n}$ with $\mathbf{A}_{G_{u,v}}=1$ if and only if $(u,v)\in E_G $. In our analysis, we consider simple, finite and undirected graphs. We will call ${\cal G}$ the set of all graphs. 
\end{definition}
 %\FF{To be consistent with the next definition better to use $u$ and $v$ also here? (Instead of $i$ and $j$)}

% \begin{definition}[\textit{neighborhood}]\label{def:neig}
% Let $G = (V_G, E_G)$ be a graph. The \textbf{neighborhood} of a node $v \in V_G$ is defined as $N(v) = \{ u \in V_G \mid (v, u) \in E_G \}$,
% The set of $m$-hop neighbors of $v$, denoted $N^m(v)$, includes all nodes at graph distance exactly $m$ from $v$; finally given two nodes $u, v \in V_G$, their $m$-hop neighborhood is defined as $N^m(u, v) = N^m(u) \cup N^m(v)$.
% \end{definition}

\begin{definition}[\textit{neighborhood}]\label{def:neig}
Let $G = (V_G, E_G)$ be a graph. Given two nodes $u, v \in V_G$, $u$ is said to be $m$-far from $v$ if there exists a path of $m$ edges connecting $u$ and $v$. Given a node $v \in V_G$, we denote by $N^m(v)$ the set of nodes $m$-far away from $v$. In particular, the 1-hop neighborhood of $v$, denoted $N(v)$, corresponds to $N^1(v)$ and includes all nodes connected to $v$, i.e., $N(v) = \{ u \in V_G \mid (v, u) \in E_G \}$. Finally, given two nodes $u, v \in V_G$, their joint $m$-hop neighborhood is defined as $N^m(u, v) = N^m(u) \cup N^m(v)$.
\end{definition}

\begin{definition}[\textit{link representation model}]\label{LRmodel}
A \textbf{link representation model} $M$ is a class of functions $F$ mapping node pairs in graphs to vector representations:
\begin{equation}
F : ((u,v),G, \mathbf{X}^0) \mapsto \mathbf{x}_{(u,v)}\in \mathbb{R}^d  
\end{equation}
with $G\in {\cal G}$, $\mathbf{X}^0\in \mathbb{R}^{n\times f}$ node feature matrix and $u,v\in V_G$.

In this work, we adopt a broad notion of \textit{link}, referring to any pair of nodes \((u,v) \in V_G \times V_G\), regardless of whether an edge between them actually exists in \(G\), i.e., whether $(u,v) \in E_G$. Accordingly, the link representation model computes a vector representation for arbitrary node pairs, not only for existing edges. Throughout the paper, the term \textit{link} will always refer to a generic node pair.

%\end{definition}
%\begin{definition}[link representation model]\label{LRmodel}
%Let $G=(V ,E ,\mathbf{X} ^0)$ be a graph. A \textbf{link representation model} is a model which aims at learning a function:
%\begin{align}
%F : (V \times V, G) &\rightarrow \mathbb{R}^d  \\
%((u, v), G) &\mapsto \mathbf{x}_{(u,v)} \in \mathbb{R}^d.
%\end{align}

\end{definition}

%Several approaches for link representation exist, ranging from heuristic methods like Adamic-Adar (AA) \citep{adamic2003friends} and Resource Allocation (RA) \citep{zhou2009predicting}, to embedding methods like Node2Vec \citep{grover2016node2vec}, to message-passing based approaches such as Graph Autoencoders (GAE)\citep{kipf2016variational} and SEAL \cite{zhang2021labeling}. .

Models that compute link representations can be applied to a variety of downstream tasks. For instance, they can be used for \textit{link prediction}~\cite{zhang2018link,lu2011link,zhou2021progresses}, where the goal is to predict the existence of an edge between two nodes, for \textit{link classification}~\cite{rossi2021knowledge,wang2021apan,cheng2025edge}, where different types of relationships are inferred and for \textit{link regression}~\cite{liang2025line,dong2019link}, where real-valued properties associated with node pairs are estimated. In this work, we focus on message-passing (MP) link representation models, which have demonstrated superior performance and robustness across a wide range of benchmarks \citep{li2023evaluating}.

\begin{definition}[\textit{GNN}]\label{gnn}
Let  $G = (V_G, E_G)$ be a graph with feature matrix $\mathbf{X}^0$. A \textbf{GNN} iteratively updates the representation of nodes $v\in V_G$ following the propagation scheme:  
\begin{align}
\mathbf{x}_v^0 = \mathbf{X}_{[v,:]}^0 \quad \quad
\mathbf{x}^l_v=\text{\small UPDATE}\left(\mathbf{x}_v^{l-1}, \text{\small AGGREGATE}\left(\{\mathbf{x}_u^{l-1} \mid u \in N(v)\}\right)\right).
\end{align}
We denote the final representation at the $L$-th layer as $\mathbf{x}_v^L= \rho(v,G,\mathbf{X}^0)$.
\end{definition}

In the rest of this paper, whenever we refer to MP models, we refer to models that employ the iterative mechanism defined in Definition \ref{gnn} to compute node representations.

A key theoretical question concerns GNNs expressive power, their ability to distinguish non-isomorphic inputs. Standard GNNs have been shown to be at most as powerful as the 1-WL test~\citep{morris2019weisfeiler,xu2019powerful}, a heuristic for graph isomorphism based on iterative multiset aggregation~\citep{weisfeiler1968reduction}. To go beyond this limit, higher-order GNNs~\citep{morris2019weisfeiler} aggregate over $k$-tuples of nodes, reaching the expressivity of the $k$-WL test; we denote by $k_\rho$ the smallest such $k$ such that $\rho$ is as powerful as $k$-WL, i.e, $k_\rho$ expresses the representational power of $\rho$ within the $k$-WL paradigm. In contrast to graph-level tasks, the expressiveness of message-passing models for link representation remains less explored~\citep{zhang2021labeling,Srinivasan2020On}. We now provide formal definitions on what does it mean for two links to be different in a graph.

\begin{definition}[\textit{node permutation}]\label{permutation}
A \textbf{node permutation} $\pi:\{1,\ldots,n\}\rightarrow \{1,\ldots,n\}$ is a bijective function that assigns a new index to each node of the graph. All the $n!$ possible node permutations constitute the permutation group $\Pi_n$. Given a subset of nodes $S\subseteq V_G $, we define the permutation $\pi$ on $S$ as $\pi(S):=\{\pi(i)|i\in S\}$. Additionally, we define $\pi (\mathbf{A}_G )$ as the matrix $\mathbf{A}_G $ with rows and columns permutated based on $\pi$, i.e., $\pi(\mathbf{A}_G )_{\pi(i), \pi(j)}=(\mathbf{A}_{G})_{{i,j}}$.
\end{definition}

\begin{definition}[\textit{automorphism}]\label{automorphism}
An $\textbf{automorphism}$ on the graph $G=(V_G ,E_G)$ is a permutation $\sigma\in \Pi_n$ such that $\sigma(\mathbf{A}_G )=\mathbf{A}_G$. All the possible automorphisms on a graph $G$ constitute the automorphism group $\Sigma_n^G$. 
\end{definition}
%Generally, not all node permutations are also automorphisms. This is exclusively true for complete graphs and empty graphs, where every permutation of nodes maintains the graph's adjacency relationships and so $\Pi_n$ is isomorphic to the symmetry group $\Sigma_n^G$.

\begin{definition}[\textit{automorphic links}]\label{isomlink}
Let  $G = (V_G, E_G)$  be a graph and $\Sigma_n^G$ its automorphism group. Two pairs of nodes  $(u, v), (u', v') \in V_G \times V_G$  are said to be \textbf{automorphic links} ( $(u, v) \simeq (u', v')$ ) if exists $ \sigma \in \Sigma_n^G$ such that $\sigma(\{u, v\}) = \{u', v'\}$.

\end{definition}

The notion of automorphic links provides the foundation for evaluating models expressivity, as an expressive model should be able to distinguish between non automorphic links.
\begin{definition}[\textit{more expressive}]\label{def:moreex}
Let \( M_1 \) and \( M_2 \) be two link representation models (Def. \ref{LRmodel}). \( M_2 \) is \textbf{more expressive} than \( M_1 \) (\( M_1 \leq M_2 \)) if, for any graph \( G = (V_G, E_G) \) and any pair \( (u, v), (u', v') \in V_G \times V_G \) with \( (u, v) \not\simeq (u', v') \):  
\begin{equation*}
\resizebox{\textwidth}{!}{$
 \exists F_1\in M_1: F_1((u, v), G, \mathbf{X}^0) \neq F_1((u', v'), G, \mathbf{X}^0) \Rightarrow \exists F_2\in M_2: F_2((u, v), G, \mathbf{X}^0) \neq F_2((u', v'), G, \mathbf{X}^0).$}    
\end{equation*}

\end{definition}
% \begin{definition}[more expressive]
% Let \( F_1 \) and \( F_2 \) be two link representation models (Def. \ref{LRmodel}). \( F_2 \) is \textbf{more expressive} than \( F_1 \) (\( F_1 \leq F_2 \)) if, for any graph \( G = (V, E, \mathbf{X}^0) \) and any pair \( (u, v), (u', v') \in V \times V \) such that \( (u, v) \not\simeq (u', v') \),  
% \begin{equation}
%  F_1((u, v), G) \neq F_1((u', v'), G) \Rightarrow F_2((u, v), G) \neq F_2((u', v'), G).    
% \end{equation}

% \end{definition}

\section{The Expressive Power of MP-based Link Representation Methods}

% \AP{I would start by explaining what this section contributes, and then showing that existing methods fit the framework.. you can then quickly introduce the subsections..}

% In this section, we describe the main state-of-the-art MP-based methods for link representation, namely Pure GNN \citep{kipf2017semisupervised, kipf2016variational, veličković2018graph, NIPS2017_5dd9db5e, xu2018how}, NCN and NCNC \citep{wangneural}, ELPH and BUDDY \citep{chamberlaingraph}, Neo-GNN \citep{yun2021neo}, SEAL \citep{zhang2021labeling} and PEG \citep{wangequivariant}. All these methods are designed to produce a link representation \FF{(Is it correct to say that the PureGNNs are "designed" to produce link representation. Maybe better "are able to"?)}, which can then be used for a variety of downstream tasks, the most common being link prediction. However, in this section, we focus exclusively on the ability of the models to generate informative link representations, independently of the specific downstream task for which the representations might be employed.
% After describing the existing methods, we introduce a general framework that encompasses all existing models, where each model can be seen as a specific instance of the framework. Finally, we leverage this framework to study the expressiveness of the models and to construct a hierarchy based on their representational power.

We propose a general framework for message-passing models that compute link representations, encompassing a broad class of existing methods as specific instances. By formalizing these models under a unified perspective, we enable a principled analysis of their expressive power. Leveraging this framework, we further establish an expressiveness-based hierarchy of existing approaches.

\subsection{MP-based methods for link representation}\label{subsec:methods}

Before defining the framework, we first review representative MP-based models, including Pure GNNs~\citep{kipf2017semisupervised,kipf2016variational,veličković2018graph,NIPS2017_5dd9db5e,xu2018how}, NCN~\citep{wangneural}, NCNC~\citep{wangneural}, ELPH~\citep{chamberlaingraph}, BUDDY~\citep{chamberlaingraph}, Neo-GNN~\citep{yun2021neo} and  SEAL~\citep{zhang2021labeling}.%, and PEG~\citep{wangequivariant}. 

\paragraph{Pure GNN Methods.}Pure GNN Methods learn representation $\mathbf{x}_{(u,v)}\in \mathbb{R}^d$ for each pair of node $(u,v)$ with $u,v\in V_G $ as $\mathbf{x}_{(u,v)}=g(\mathbf{x}_u^L, \mathbf{x}_v^L)$, where $g$ is an aggregation function and $\mathbf{x}_u^L, \mathbf{x}_v^L$ are the node representation of $u$ and $v$ learned by the GNN at the final layer $L$. %The link representation can then be used for several downstream tasks, like for example, predicting the probability of existence of the link. 
In principle, any type of GNN can be used, and the function $g$ can be modeled using an MLP over any aggregation function over the feature vectors. In practice, the most commonly used pure GNN model is GAE~\citep{kipf2016variational}. %This method employs a GCN~\citep{kipf2017semisupervised} to generate node representations. These representations are then aggregated using the inner product, followed by the application of a sigmoid function to produce the final probability prediction.\\ 

\paragraph{Neo-GNN.}
%Neo-GNN is a graph neural network designed for link prediction that explicitly incorporates structural information from the adjacency matrix. Instead of relying solely on node features, Neo-GNN learns structural features that capture neighborhood overlap, a key heuristic for link prediction. 
Given a graph $G=(V_G ,E_G)$, Neo-GNN computes a node representation matrix $\mathbf{Z}$ as: 
\begin{equation}
\mathbf{Z} = \sum_{l=1}^{L} \beta^{l-1} \mathbf{A}^l_G \mathbf{X}^{\text{struct}} \quad \text{with} \quad \mathbf{X}^{\text{struct}} = \text{diag}(\mathbf{x}^{\text{struct}}), \; \mathbf{x}_v^{\text{struct}} = \mathcal{F}_\Theta((\mathbf{A}_G)_v)
\label{eq:neo}
\end{equation}
where $\mathcal{F}_\Theta$ is a learnable function over the adjacency matrix $A_G$, $L$ is the maximum hop considered, and $\beta\in (0,1)$ controls the weight of distant neighbors. The final link representation is computed as an aggregation of the node representation obtained through standard GNN and the one in $\mathbf{Z}$, namely $ \mathbf{x}_{(u,v)}= g((\mathbf{z}_u,\mathbf{z}_v),(\mathbf{x}_u^L, \mathbf{x}_v^L))$.

%where $\alpha$ is a learnable parameter balancing structural and feature-based predictions and $s$ is an aggregation function. Neo-GNN generalizes neighborhood overlap heuristics (e.g., CN and AA).

\paragraph{NCN.}

NCN computes the representation of node pairs using the representations of their common neighbors. NCN defines the link representation as:
\begin{equation}\label{eq:ncn}
\mathbf{x}_{(u,v)} = \mathbf{x}_u^L \odot \mathbf{x}^L_v \, \Vert \sum_{i \in N(u) \cap N(v)} \mathbf{x}^L_i,
\end{equation}
where $\mathbf{x}_u^L$ is the node representation at the final $L-$th layer of a GNN. % $\odot$ denotes element-wise multiplication, $\Vert$ represents concatenation, and $N(u) \cap N(v)$ is the set of common neighbors of nodes $u$ and $v$. 
NCNC extends NCN to deal with settings with incomplete topological information (e.g., for link prediction or graph completion), by considering not only nodes in $N(u)\cap N(v)$ but all nodes in $N(u)\cup N(v)$ and calculating for them the probability of being common neighbors with NCN. 

% to address the issue of missing edges by predicting potential missing links and updating the common neighbor set accordingly. The resulting link representation is:
% \begin{equation}\label{eq:ncnc}
% \mathbf{x}_{(u,v)} = \mathbf{x}_u^L \odot \mathbf{x}_v^L \, \Vert \sum_{i \in N(u) \cup N(v)}  P_{uvi} \mathbf{x}_i^L,
% \end{equation}
% where \(P_{uvi}\) is equal to one if and only if \(i \in N(u) \cap N(v)\); if \(i \in N(u) \setminus N(v)\), \(P_{uvi}\) is set to the probability \(p(i,v)\) predicted by NCN; and if \(i \in N(v) \setminus N(u)\), \(P_{uvi}\) is set to the probability \(p(i,u)\) predicted by NCN.

\paragraph{SEAL.}
%SEAL is a GNN-based method for link prediction that can learn structural link representations by extracting and labeling enclosing subgraphs around the link to be predicted. 
Given a graph $G=(V_G ,E_G)$, for each link $(u, v)$, SEAL constructs an $h$-hop enclosing subgraph $G_{uv}^{(h)}$, which contains all nodes within $h$ hops of $u$ and $v$. Nodes in $G_{uv}^{(h)}$ are labeled using the Double-Radius Node Labeling (DRNL)\citep{zhang2021labeling}, which assigns different labels based on distances to $u$ and $v$. The representation of link $(u,v)$ is obtained processing the graph $G_{uv}^{(h)}$ with a GNN. %It has been proved in \cite{zhang2021labeling} that SEAL can learn a structural link representation. %\MJ{One must be careful with the usage of 'learns' vs. 'is able to learn' or 'can represent'}

\paragraph{ELPH.}
In ELPH, two types of count-based features are computed: (i) \( A_{uv}[d_u, d_v] \), i.e., the number of nodes at distance \( d_u \) from node \( u \) and \( d_v \) from node \( v \); and (ii) \( B_{uv}[d] \), the number of nodes at distance \( d \) from \( u \) and more distant than $d$ from \( v \).  These counts are estimated using \textit{MinHash} (for Jaccard similarity) and \textit{HyperLogLog} (for set cardinality). 
% These features are used as edge features in the message passing layer:
% \begin{align}
% \mathbf{x}_{(u,v)}^l&=\{A_{uv}[d_u, l],A_{uv}[l, d_v], B_{uv}[l],B_{vu}[l]\mid d_u,d_v<l\}\\
% \mathbf{x}_v^l&= \text{\small UPDATE}(\mathbf{x}_v^{l-1}, \text{\small AGGREGATE}(\{(\mathbf{x}_v^{l-1}, \mathbf{x}_u^{l-1},\mathbf{x}_{(u,v)}^l )\mid u \in N(v)\})
% \end{align}
The link representation is obtained as:
\begin{equation}
\mathbf{x}_{(u, v)} = g( \mathbf{x}^{L}_u, \mathbf{x}^{L}_v,\ \{ {A}_{uv}[d_u,d_v],{B}_{uv}[d] \mid d,d_u,d_v<[L]\}),   
\end{equation}
where $g$ in an aggregation function and $L$ is the final layer. To overcome ELPH's memory limitations, BUDDY precomputes structural features via offline graph traversals and sketching. %avoiding online message passing.

% \paragraph{PEG}
% The PEG layer incorporates positional encodings computed as the first \( p \) eigenvectors of the normalized graph Laplacian. Message passing is modified by weighting edges based on distances in the PE space:
% \begin{equation}
%     \mathbf{x}^l_v=\text{\small UPDATE}\left(\mathbf{x}_v^{l-1}, \text{\small AGGREGATE}\left(\{\mathbf{x}_u^{l-1}\cdot \phi(\|\mathbf{z}_u - \mathbf{z}_v\|) \mid u \in N(v)\}\right)\right)
% \end{equation}
% where \( \mathbf{Z} \in \mathbb{R}^{n\times p} \) contains the first Laplacian eigenvectors. The representation of link $(u,v)$ is then calculated as $\mathbf{x}_{(u,v)}=(\mathbf{x}^L_v, \mathbf{x}^L_v, \mathbf{z}_u^T\mathbf{z}_v) $ 

\subsection{A General Framework for MP-based Link Representation Methods}\label{subsec:kml}

\begin{table}[b]
\caption{Model formulations from Section \ref{subsec:methods} expressed within the $k_{\phi}$-$k_{\rho}$-$m$ framework. A ‘/’ indicates that the corresponding component is not included in the model.}\label{tab:fram}
\centering
\resizebox{\textwidth}{!}{%
\begin{tabular}{lcccccccc}
\toprule
\textbf{Model} & COMB & $g$ & $k_{\phi}$ & AGG & $\psi$ & $k_{\rho}$ & $m$ & $f$ \\
\midrule
\rowcolor{verylightgray}Pure GNN & / & $\odot$ & 1-WL & / & / & / & / & / \\
\rule{0pt}{15pt}NCN & $\parallel$ & $\odot$ & 1-WL & $\sum$ & $\rho\left(i;G,\mathbf{X}^0\right)$ & 1-WL & 1 & $\mathbf{X}^0$ \\

%\rule{0pt}{15pt}NCNC & $\parallel$ & $\odot$ & 1-WL & $\sum$ & $\alpha\cdot\rho\left(i;G,\mathbf{X}^0\right)\quad\alpha\in(0,1)$ & 1-WL & 2 & $\mathbf{X}^0$ \\

\rowcolor{verylightgray}\rule{0pt}{15pt}ELPH & $\parallel$ & $\parallel$ & 1-WL & $\sum$ & $\rho\left(i;G,\mathbf{X}^1\right)\cdot\prod\limits_{r=1}^{m}\prod\limits_{d=1}^{m}\mathds{1}_{dr}(i)$ & 1-WL & m & $\mathbf{x}^1_i=1$ \\

\multirow{2}{*}{Neo-GNN} & \multirow{2}{*}{$\parallel$} & \multirow{2}{*}{$\parallel$} & \multirow{2}{*}{1-WL} & \multirow{2}{*}{$\sum$} & \rule{0pt}{15pt}$b\cdot\rho\left(i;G,\mathbf{X}^0\right)$ \quad with & \multirow{2}{*}{1-WL} & \multirow{2}{*}{m} & \multirow{2}{*}{$\mathbf{X}^0$} \\
& & & & &\rule{0pt}{15pt}$b=\sum\limits_{r=1}^{m}\sum\limits_{d=1}^{m}A_{uv}^r\cdot A_{uv}^d$ & & & \\
\rowcolor{verylightgray}\rule{0pt}{15pt}SEAL & / & / & / & $\sum$ & $\rho\left(i;G,\mathbf{X}^D\right)$ & 1-$|N^m(u,v)|$-WL & m & $\mathbf{x}_i^D = \mathbf{x}^0_i\parallel\min\limits_{u,v}(\delta(i,u),\delta(i,v)) + 1$ \\
% \rule{0pt}{15pt}PEG & $\parallel$ & $\parallel$ & $>\text{WL}$ & $\langle\cdot\rangle$ & $\mathbf{z}_u^F\cdot \mathbf{z}_v^F$, $\mathbf{Z}$ s.t. $\mathbf{L}_G = \mathbf{Z}\Lambda \mathbf{Z}$ & $\infty\text{-WL}$ & \textcolor{red}{?} & / \\
\bottomrule
\end{tabular}}
\end{table}

Building on the definitions introduced in Section~\ref{subsec:methods}, we observe that MP-based approaches for link representation generally follow a common paradigm: they combine the representations of the two endpoint nodes with additional pairwise information extracted from their local neighborhoods, except for pure GNN methods, which rely solely on node representations. Although specific mechanisms and architectural designs vary, the core structure of these models can be systematically characterized by two key factors: (i) the link neighborhood radius, which defines the portion of the graph considered when extracting pairwise information for the link, and (ii) the expressiveness of the message-passing (MP) functions used to compute the node representations. Importantly, the link neighborhood radius is distinct from the radius induced by the depth of the MP function itself. Motivated by these common factors, we introduce a general framework that unifies existing approaches under a common formalism. In this framework, the link representation is constructed by combining: (1) the representations of the two endpoint nodes, and (2) a representation of the link’s $m$-order neighborhood, each computed via MP functions that may have different levels of expressive power and different number of layers.
We formalize this in the following definition, which we refer to as the \textit{$k_\phi$-$k_\rho$-$m$ framework} for link representation. For ease of readability, we omit the graph $G \in \mathcal{G}$ as an explicit argument in message-passing and link-representation functions, assuming it is implicitly defined as the underlying structure on which these functions operate.

\begin{definition}[\textit{$k_\phi$-$k_\rho$-$m$ framework} ]\label{def:kml}
Given a graph $G=(V_G,E_G)$ with feature matrix $\mathbf{X}^0$, a MP link representation model $M$ belongs to the $k_\phi$-$k_\rho$-$m$ framework if its functions can be expressed as:
{\footnotesize
\begin{align}
F((u, v), \mathbf{X}^0) = \text{\small COMB}\Big(
g\big(\phi(u, \mathbf{X}^0), &\phi(v, \mathbf{X}^0)\big),\text{\small AGG}\big(\{f(i,u,v,\mathbf{X}^0)\mid i \in \bigcup_{j=0}^{m} N^j(u,v)\}\big)
\Big)\\
f(i,u,v,\mathbf{X}^0) &= \psi\big(\rho(i,h(u,v,\mathbf{X}^0)),u,v\big)
\end{align}
}

\begin{figure}[b]
    \centering
    \includegraphics[width=0.95\linewidth]{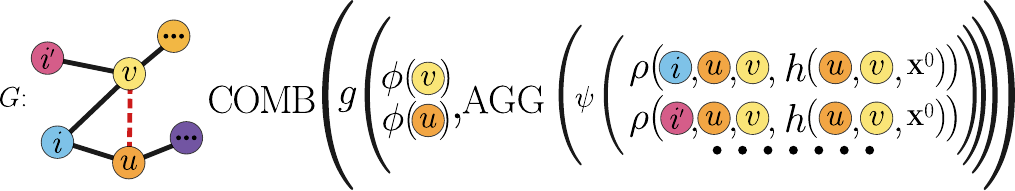}
    \caption{$k_\phi$-$k_\rho$-$m$ framework: nodes representations in the m-order neighborhood of target link are calculated using a MP function $\rho$ with possibly modified node features; these representations are aggregated and combined with representations of endpoints of target link obtained through MP function $\phi$.}
    \label{fig:fram}
\end{figure}

where $\phi$ and $\rho$ are MP functions (Definition \ref{gnn}) with expressive power respectively of $k_{\phi}$ and  $k_{\rho}$, $h(u,v,G,\mathbf{X}^0)\in \mathbb{R}^{n\times d}$ is a new node feature matrix computed using the original feature matrix and possibly pairwise information, $\psi$ scales the message passing representations by a coefficient incorporating pairwise information from the graph, $\text{\small AGG}$ is an aggregation function over the representations of nodes in $\bigcup_{j=0}^{m} N^j(u,v)$ and $\text{\small COMB}$ combine the endpoints representations with the neighborhood representation. %We will indicate the set of link representation models expressible bu the $k_{\phi}$-$k_{\rho}$-$m$ framework as $\mathcal{F}$.
\end{definition}

Table \ref{tab:fram} provides the formulation of all the models presented in Section \ref{subsec:methods} using the $k_{\phi}$-$k_{\rho}$-$m$ framework, while Figure \ref{fig:fram} provides a visualization of the framework components.
Through the $k_\phi$-$k_\rho$-$m$ framework, we can systematically study the expressive power of MP-based link representation methods.  %\AP{Connect the theorem to the sentence to introduce it.. also it would be nice to give (semantically meaningful) names to theorems}
Let $\mathcal{M}$ be the set of link representation models expressible by the $k_{\phi}$-$k_{\rho}$-$m$ framework. Let $M\in \mathcal{M}$, we indicate with $k_\phi^M$-$k_\rho^M$-$m^M$ the specific value that $k_\phi$-$k_\rho$-$m$ assume in the model $M$.
% \begin{theorem}\label{theo:fram}
% Let $\mathcal{F}$ be the set of link representation models expressible by the $k_{\phi}$-$k_{\rho}$-$m$ framework. Let $M\in \mathcal{F}$, the following hold:
% \begin{enumerate}[itemsep=2pt, topsep=2pt]
% \item Let $F\in M$ with $m=0$, then, regardless of $k_\phi$, $F$ is not able to distinguish between links whose endpoints are automorphic, i.e., 
% \begin{align}
% \forall (u, v), (u', v') \text{ s.t. } \exists \sigma_1, \sigma_2 \in \Sigma_n^G &\text{ with } \sigma_1(u) = u' \land \sigma_2(v) = v'\\
% F((u', v'), \mathbf{X}^0) &= F((u, v),\mathbf{X}^0)
% \end{align}
% therefore, automorphic links whose endpoints are automorphic will be assigned to the same representation. Moreover, given $F_1,F_2 \in \mathcal{F}$, if $F_1$ uses just $g_1$ and $\phi_1$ while $F_2$ uses also $f_2$ with an injective $\text{\small COMB}$, regardless of $k_{\phi_1}$ and $k_{\phi_2}$ , $F_1 \leq F_2$.
% \item Let $F_1,F_2 \in \mathcal{F}$ with $k_{{\rho}_1}=k_{{\rho}_2}=\textrm{1-WL}$ and let $l_1,l_2$ be the number of layers used respectively by $F_1,F_2$. If $m_1+l_1 \leq m_2+l_2$, then $F_1\leq F_2$.
% \item Let $F_1,F_2 \in \mathcal{F}$ with $m_1 = m_2$. If $k_{\rho_1} \leq k_{\rho_2}$, then $F_1 \leq F_2$.
% \end{enumerate}
% \end{theorem}

\begin{theorem}\label{theo:fram}
Let $M\in \mathcal{M}$, the following hold:
\begin{enumerate}[itemsep=2pt, topsep=2pt]
\item If $m^M=0$, then, regardless of $k_\phi^M$, $M$ is not able to distinguish between links whose endpoints are automorphic, i.e., 
\begin{align}
\forall F\in M, \forall (u, v), (u', v') \text{ s.t. } \exists \sigma_1, \sigma_2 &\in \Sigma_n^G \text{ with } \sigma_1(u) = u' \land \sigma_2(v) = v'\\
F((u', v'), \mathbf{X}^0) &= F((u, v),\mathbf{X}^0)
\end{align}
therefore, automorphic links whose endpoints are automorphic will be assigned to the same representation. Moreover, given $M_1,M_2 \in \mathcal{M}$, if $m^{M_1}=0$, while $m^{M_2}>0$ and $\text{\small COMB}$ is injective, regardless of $k^{M_1}_{\phi}$ and $k^{M_2}_{\phi}$ , $M_1 \leq M_2$.
\item Let $M_1,M_2 \in \mathcal{M}$ with $k^{M_1}_{{\rho}}=k^{M_2}_{{\rho}}=\textrm{1-WL}$ and let $l^{M_1},l^{M_2}$ be the number of layers used respectively by $M_1,M_2$. If $m^{M_1}+l^{M_1} \leq m^{M_2}+l^{M_2}$, then $M_1\leq M_2$.
\item Let $M_1,M_2 \in \mathcal{M}$ with $m^{M_1} = m^{M_2}$. If $k^{M_1}_{\rho} \leq k^{M_2}_{\rho}$, then $M_1 \leq M_2$.
\end{enumerate}
\end{theorem}

Proof is provided in Appendix~\ref{app:proofs}. Intuitively, the theorem highlights that relying solely on the representations of the endpoints through $g$ and $\phi$ is fundamentally limited: even if $\phi$ is highly expressive, such models fail to distinguish links whose endpoints are automorphic. This limitation arises from the fact that the $k$-WL algorithm, preserves graph automorphisms for every $k$ \citep{lichter2025computational,dawar2020generalizations,cai1992optimal}. Instead, a more efficient and expressive modeling strategy is to compute $\rho$ over a suitable $m$-order neighborhood using an appropriate number of MP layers. In many models, the number of such layers is a design choice: for instance, ELPH encodes link information by counting nodes at various distances from the link, which effectively corresponds to zero message-passing layers. Conversely, in Neo-GNN, node representations are computed through a function on the adjacency matrix, effectively implementing a single round of message passing. This distinction is crucial: even models with large neighborhoods $m$ but few message-passing layers can be outperformed—both in expressiveness and efficiency—by models with smaller $m$ but deeper architectures. Thus, controlling expressiveness via the number of layers offers a more principled and computationally efficient path. Finally, increasing the expressiveness of the base model $\rho$ on the $m$-order neighborhood leads to more expressive representations. One example of model with more expressive $\rho$ is SEAL which increases expressiveness by adding link-aware positional encodings to node features, effectively simulating the power of $1$-$|N^m(u,v)|$-WL, which is strictly more expressive than 1-WL \citep{zhou2023relational}.
However, increasing expressivity beyond $1$-WL comes at a computational cost: for example, the complexity of $k$-WL is $O(n^{k+1}\log n)$~\citep{immerman2019k}.

Using Theorem \ref{theo:fram} it is possible to study the expressiveness of existing methods. The following Theorem introduces a hierarchy of existing methods.

\begin{theorem}\label{theo:model}
%Let $m_{\text{model}}$ and $l_{\text{model}}$ be the radius and the number of layers used by $\text{model}\in \{\text{NCN, Neo-GNN, ELPH, SEAL}\}$. 
The following hold:
\begin{enumerate}[itemsep=2pt, topsep=2pt]
    \item For any radius and number of layers, NCN, Neo-GNN, ELPH, SEAL are more expressive than Pure GNNs;
    \item if $l^{NCN}\geq m^{Neo-GNN}$, NCN is more expressive then Neo-GNN;
    \item if $m^{Neo-GNN}\geq m^{ELPH}-1$, Neo-GNN is more expressive than ELPH;
    \item if $l^{NCN}\geq m^{ELPH}-1$, NCN is more expressive then ELPH;
    \item SEAL is more expressive then NCNC, NCN, ELPH, Neo-GNN.
\end{enumerate}

\end{theorem}

Proof is provided in Appendix~\ref{app:proofs} and rely on Theorem \ref{theo:fram} and the formal expression of models within the theoretical framework described in Table \ref{tab:fram}. 

The $k_\phi$-$k_\rho$-$m$ framework serves as a foundation for analyzing the expressiveness of link representation models in a principled and structured way. In the next section, we complement this theoretical perspective by introducing a synthetic evaluation protocol designed to empirically assess the expressiveness of link representation models.

\section{A Synthetic Evaluation Procedure for Link Representation Expressiveness}
\label{sec:synthetic}

We introduce a synthetic protocol to empirically assess a MP model’s ability to assign distinct representations to structurally different links. The protocol is designed to help evaluate new link representation models in a simple and controlled setting. It consists of a synthetic dataset and an evaluation framework. The synthetic dataset and the code are publicly available\footnote{\url{https://anonymous.4open.science/r/link-representation-gnn-8124/README.md}}.

\subsection{\mbox{\texttt{LR-EXP}: A Synthetic Benchmark for Link-Level} Expressiveness}
%\subsection{\texttt{LR-EXP}: A Synthetic Benchmark for Link-Level Expressiveness}

While several synthetic datasets have been proposed to evaluate the expressive power of GNNs in distinguishing non-isomorphic graphs \citep{bianchi2023expressive,wang2023empirical,abboud2020surprising,murphy2019relational,balcilar2021breaking,bouritsas2022improving}, a comparable benchmark for assessing link-level expressiveness is still missing, despite its potential utility as a fast and controlled way to empirically assess the expressivity of both existing and new models. To address this gap, we propose a novel synthetic dataset, \SyntData, aimed at measuring a model’s ability to assign different outputs to non-automorphic links. To make the distinction task challenging, the dataset includes non-automorphic links whose endpoint nodes share the same 1-WL colors. This is a critical design choice: non-automorphic links with differently colored nodes are trivially distinguishable by any pure GNN with an injective aggregation function $g$. \SyntData includes 1400 graphs, each containing at least one such pair of non-automorphic links. 
\begin{wrapfigure}{r}{0.34\textwidth}
    \centering
    \includegraphics[width=0.33\textwidth]{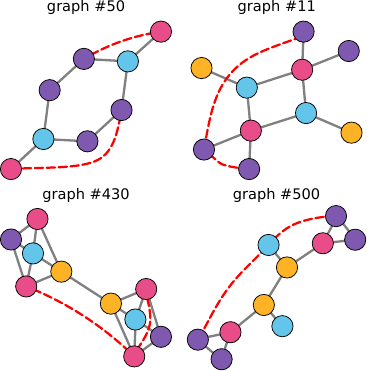}
    \caption{Examples of \SyntData graphs. Node colors show WL colors; dashed lines mark test links that are non-automorphic but indistinguishable for standard GNNs.}
    \label{fig:synthetic}
\vspace{-9mm}
\end{wrapfigure}
Each graph is generated in two steps: (1) sample an Erdős–Rényi graph with $n \in \{5, \dots, 17\}$ nodes and edge probability $p \sim \mathcal{U}(0, 1)$; (2) duplicate the graph and add inter-block edges with probability $p' \sim \mathcal{U}(0, 1)$. This construction introduces rich symmetries and increases the probability of obtaining node pairs that are non-automorphic yet indistinguishable under the WL test. We retain only those graphs where such link pairs exist, and use them to define training, validation, and test sets. Figure~\ref{fig:synthetic} shows examples of these link pairs. Importantly, while \SyntData is relatively small in scale, this is by design: our goal is not to evaluate scalability, but to isolate and probe expressivity. Since expressiveness concerns a model's ability to produce different outputs for structurally distinct inputs, rather than generalization over large data volumes, dataset size is not a confounding factor for the phenomenon we aim to study.

\subsection{Evaluation Framework}
\label{sec:evaluationsynthetic}

Our training and evaluation approach is inspired by the method proposed in \citet{wang2024an}. In particular, we employ a siamese network design \cite{bromley1993signature}, with
central component consisting of two identical models with identical parameters, each processing one link independently to generate separate embeddings. For a pair of links $(u,v)\not\simeq(u',v')\in V_G\times V_G$, and a model $F$, the model generates a corresponding pair of embeddings. The model is trained with the goal of encouraging separation of links embeddings using the loss:
\begin{equation}
 L(F, (u,v),(u',v'), \mathbf{X}^0) = \max \left( 0, \frac{F((u,v), \mathbf{X}^0) \cdot F((u',v'), \mathbf{X}^0)}{\|F((u,v), \mathbf{X}^0)\| \|F((u',v'), \mathbf{X}^0)\|} \right),   
\end{equation} 
Training plays a crucial role in realizing the theoretical expressiveness, by exploring the function space, enforcing properties like injectivity, and reducing numerical errors.

To evaluate models, we compare their outputs on pairs of non-automorphic links. If the resulting representations differ sufficiently, this indicates successful discrimination. However, selecting an appropriate threshold is challenging: a high threshold may cause false negatives, where true distinctions fall below the threshold; a low threshold may cause false positives, where random noise creates apparent differences. To address this, we adopt the Reliable Paired Comparison method~\cite{wang2024an}, which assesses groups of results using two components: the Major Procedure, quantifying representation differences, and the Reliability Check, measuring internal variability. A pair is considered distinguishable only if both checks are passed. As all pairs are non-automorphic, the task is framed as one-class classification, and we report precision as the fraction of correctly distinguished pairs. See Appendix~\ref{app:synthetic_setting} for details.

\subsection{Experimental results on \SyntData}

We evaluated the expressive power of all models from Section \ref{subsec:methods} on \SyntData. As discussed in Section~\ref{subsec:kml}, some models treat $m$ as a fixed design choice while keeping the number of layers $l$ as a tunable parameter, and vice versa. Even in models where $m$ is left as a free parameter, practical implementations impose constraints due to computational costs. For example, both ELPH and SEAL limit $m$ to a maximum of 3. In our experiments, we adopt $m = 3$ for all models that allow it, ensuring a fair comparison. We set the number of layers $l = 3$ across all methods, which strikes a balance between expressiveness and avoiding oversmoothing effects \citep{cai2020note}. Under this configuration, Theorem \ref{theo:model} shows that NCN is more expressive than ELPH and matches the expressiveness of Neo-GNN, while SEAL is the most expressive method. Details of the implementation are provided in Appendix~\ref{app:experimental-details}. The experimental results on \SyntData provide empirical validation of our theoretical framework. Table \ref{tab:results} shows that Pure GNN methods (\GCN, \GAT, \GAE, \SAGE) completely fail to distinguish non-automorphic links, achieving zero precision. This is a consequence of their fundamental limitation - the automorphic node problem we analyzed in Section~\ref{subsec:kml}, which prevents them from generating distinct representations for links composed by automorphic nodes. 
\begin{wraptable}{l}{0.35\textwidth}\label{tab:results}
\vspace{-4mm}
\caption{Test precision (mean $\pm$ std over 5 runs) for different models. Higher model expressiveness leads to higher test precision.}
    \centering
    \setlength{\tabcolsep}{4pt}
    \begin{tabular}{l c}
        \toprule
        \textbf{Model} & \textbf{Test Precision(\%)} \\
        \midrule
        \GCN & 0 {\scriptsize ($\pm$ 0)} \\
        \GAT & 0 {\scriptsize ($\pm$ 0)} \\
 %       \GIN & 0 {\scriptsize ($\pm$ 0)} \\
        \GAE & 0 {\scriptsize ($\pm$ 0)} \\
        \SAGE & 0 {\scriptsize ($\pm$ 0)} \\
        \BUDDY & 45 {\scriptsize ($\pm$ 1)} \\
        \ELPH & 62 {\scriptsize ($\pm$ 7)} \\
        \NEOGNN & 75 {\scriptsize ($\pm$ 2)} \\
        \NCN & 75 {\scriptsize ($\pm$ 1)} \\
%        \NCNC & 30 {\scriptsize ($\pm$ 1)} \\
        \SEAL & 97 {\scriptsize ($\pm$ 0)} \\
    %    \PEG & 74 {\scriptsize ($\pm$ 0)} \\
        \bottomrule
    \end{tabular}   
\vspace{-3mm} 
\end{wraptable}
The remaining methods demonstrate progressively better performance as their expressive power increases. The performance gap between \BUDDY (45\%) and \ELPH (62\%) stems from \BUDDY's approximate counting of common neighbors (Section \ref{subsec:methods}), which reduces its expressiveness as shown in~\cite{chamberlaingraph}. \NEOGNN and \NCN\footnote{We do not report results for NCNC, as it coincides with NCN for fully observed topologies.} %on our dataset. NCNC is designed for settings with incomplete graphs, whereas in our case, observed links perfectly match the graph edges, making the probability of existence binary (1 if present, 0 otherwise) thus rendering NCNC equivalent to NCN.
 achieve comparable precision (75\%), outperforming both \ELPH (62\%) and \BUDDY (45\%). Finally, \SEAL demonstrates its high expressiveness, being able to correctly distinguish 97\% of links.

\section{Do More Expressive Models Perform Better in Real-World Benchmarks?}
\label{sec:realworld}
An interesting and practically relevant question is whether increased expressivity in link representation translates into better performance on downstream tasks involving real-world data \citep{jogl2024expressivity,morris2024position}. Among the various tasks enabled by expressive link representations, we focus on link prediction due to its broad impact across domains such as recommender systems~\citep{ying2018graph}, knowledge graph completion~\citep{nickel2015review}, and biological interaction prediction~\citep{jha2022prediction}. Intuitively, link prediction should particularly benefit from models with high expressive power when the dataset contains many non-automorphic links whose endpoint nodes are automorphic, a setting where Pure GNNs struggle. In this section, we aim to answer the following three questions: (1) How can we formally quantify the presence of non-automorphic links with automorphic endpoints in a dataset? (2) Are there real-world link prediction benchmarks where this property is especially prevalent? (3) How do models with varying levels of expressiveness perform across datasets with different levels of structural difficulty?

We address these questions by first introducing a metric to quantify the link-level symmetry of a dataset. We then apply this metric to a collection of real-world link prediction benchmarks and analyze their structural difficulty. Finally, we evaluate the performance of several state-of-the-art models with different expressive capabilities across these datasets. Code is available at\footnote{\url{https://anonymous.4open.science/r/link-representation-gnn-8124/README.md}}.

\subsection{A Metric for Measuring the Symmetry of Graphs}
We introduce a metric to quantify the presence of non automorphic links with automorphic endpoints in a dataset. This occurs when the underlying graph exhibits a high number of non-trivial automorphisms that is, intuitively, when the graph is highly symmetric. The symmetry of a graph is tightly connected to the concept of \textit{node orbits}, which we define below. 

\begin{definition}[\textit{orbit}]\label{orbit}
The \textbf{orbit} of a vertex $v\in V_G$ under the action of the automorphism group $\Sigma_n^G$ is defined as the set of nodes of $V_G$ to which $v$ can be mapped via an automorphism $\sigma\in \Sigma_n^G$. Formally, $\text{Orb}(v)= \{\sigma(v) \mid \sigma \in \Sigma_n^G\}$. We denote with $O_G= \{\text{Orb}(v)\mid v\in V_G\}$ the  set of all the orbits of $V$ with respect to the action of $\Sigma_n^G$.
% \begin{equation}
%     \text{Orb}(v)= \{\sigma(v) \mid \sigma \in \Sigma_n^G\}.
% \end{equation}
% We call $O_G= \{\text{Orb}(v)\mid v\in V_G\}$ the set of all the orbits of $V$ with respect to the action of $\Sigma_n^G$.
\end{definition}

Following~\cite{ball2018symmetric}, it is possible to define a metric that mesures how much a graph is symmetric as follows:

\begin{definition}[\textit{graph symmetry measure $r_G$}]\label{def:r}
Given a graph $G=(V_G,E_G)$, its \textbf{symmetry measure} is calculated as:
\begin{equation}\label{eq:symm_orig}
    r_G = 1 - \frac{|O_G| - 1}{|V_G| - 1}
\end{equation}
where $|O_G|$ denotes the number of orbits of the graph and $|V_G|$ is the number of nodes.   
\end{definition}
Computing all node orbits is computationally expensive. However, the number of orbits can be efficiently approximated using the number of distinct colors assigned by the WL test at convergence $WL_G$ \citep{lachi2024simple,morris2023orbit}, i.e., 
\begin{equation}\label{eq:symm_new} 
    \hat{r}_G = 1 - \frac{|WL_G| - 1}{|V_G| - 1}
\end{equation}

\begin{table}[t]
% \caption{MRR on real-world link prediction datasets sorted by increasing $\hat{r}_G$. The top three results are highlighted using \first{first}, \second{second}, and \third{third}. \texttt{OOM} means Out Of Memory, \texttt{>24h} means more than 24 hours. As symmetry increases, only more expressive models maintain top performance.}\label{tab:symm}

\caption{MRR on real-world link prediction datasets sorted by increasing $\hat{r}_G$. The top three results are highlighted using \first{first}, \second{second}, and \third{third}. \texttt{OOM} means Out Of Memory, \texttt{>24h} means more than 24 hours. As symmetry increases, only more expressive models maintain top performance. Results are averaged over 5 runs with different random seeds; standard deviations are reported in Appendix~\ref{app:results-std}.}\label{tab:symm}

\centering
\setlength{\tabcolsep}{4pt}

\resizebox{\textwidth}{!}{%
\begin{tabular}{lcccccccccccc}
& \multicolumn{12}{c}{Graph Symmetry ($\rightarrow$)}\\
\toprule
% \multirow{2}{*}{\textbf{Models}} & \textbf{ogbl} & \multirow{2}{*}{\textbf{Cora}} & \multirow{2}{*}{\textbf{Citeseer}} & \multirow{2}{*}{\textbf{Pubmed}} & \textbf{ogbl} & \textbf{ogbl} & \textbf{ogbl} & \multirow{2}{*}{\textbf{NSC}} & \multirow{2}{*}{\textbf{YST}} & \multirow{2}{*}{\textbf{GRQ}} &\multirow{2}{*}{\textbf{aifb}} & \textbf{edit}\\
%  & \textbf{citation2} &  & &  & \textbf{ddi} & \textbf{ppa} & \textbf{collab}& & & & & \textbf{tsw} \\
 % \multirow{2}{*}{\textbf{Models}} & ogbl & \multirow{2}{*}{Cora} & \multirow{2}{*}{Citeseer} & \multirow{2}{*}{Pubmed} & ogbl & ogbl & ogbl & \multirow{2}{*}{NSC} & \multirow{2}{*}{YST} & \multirow{2}{*}{GRQ} & \multirow{2}{*}{aifb} & edit \\
 % & citation2 &  &  &  & ddi & ppa & collab &  &  &  &  & tsw \\
 \multirow{2}{*}{\textbf{Models}} & \textsc{ogbl} & \multirow{2}{*}{\textsc{Cora}} & \multirow{2}{*}{\textsc{Citeseer}} & \multirow{2}{*}{\textsc{Pubmed}} & \textsc{ogbl} & \textsc{ogbl} & \textsc{ogbl} & \multirow{2}{*}{\textsc{NSC}} & \multirow{2}{*}{\textsc{YST}} & \multirow{2}{*}{\textsc{GRQ}} & \multirow{2}{*}{\textsc{aifb}} & \textsc{edit} \\
 & \textsc{citation2} &  &  &  & \textsc{ddi} & \textsc{ppa} & \textsc{collab} &  &  &  &  & \textsc{tsw} \\
\cmidrule(l{2pt}r{2pt}){2-13}
%$\hat{r}_G$ & 0.006 & 0.016 & 0.019 & 0.084 & 0.099 & 0.121 & 0.134 & 0.161 & 0,236 & 0,302 & 0,411 & 0,674 \\
$\hat{r}_G$ & 0.01 & 0.02 & 0.02 & 0.08 & 0.10 & 0.12 & 0.13 & 0.16 & 0.23 & 0.30 & 0.41 & 0.67 \\
\midrule
\GCN     & 19,98 & \second{16.61} & 21.09 & 7.13 & \first{13.46} & 26.94 & \second{6.09} & 26.79  & 1.17  & 6.89   & 10.85  & 4.12  \\
\GAT     & \texttt{OOM} & 13.84 & 19.58 & 4.95 & \second{12.92} & \texttt{OOM} & 4.18 & 2.47   & 0.40  & 0.63   & 1.07   & 0.40  \\
\SAGE    & \second{22.05} & 14.74 & 21.09 & \first{9.40} & 12.60 & 27.27 & 5.53 & 20.32 & 1.77& 1.34&0.54&8.35\\
\GAE     & \texttt{OOM} & \first{18.32} & \second{25.25} & 5.27 & 3.49 & \texttt{OOM} & \texttt{OOM}  & 16.61  & 2.32  & 2.32   & 10.98  & 3.54  \\
\BUDDY   & 19.17 & 13.71 & 22.84 & 7.56 & 12.43 & 27.70 & \third{5.67} & 18.40  & \second{13.68} & 46.23  & \second{13.73}  & \second{22.66} \\
\NEOGNN & 16.12 & 13.95 & 17.34 & \third{7.74} & 10.86 & 21.68 & 5.23 & 22.86  & 5.78  & 29.51  & 13.11  & 8.10   \\
\NCN     & \first{23.35}& 14.66 & \first{28.65} & 5.84 & \third{12.86} & \first{35.06} & 5.09  & \second{30.36}  & 11.99  & \second{48.45}  & 6.47   & \third{10.32} \\
\NCNC    & 19.61 & \third{14.98} & \third{24.10} & \second{8.58} & \texttt{>24h} & \second{33.52} & 4.73  & \third{30.20}  & \third{13.54} & \third{47.60}  & \texttt{OOM} & 10.30  \\
\SEAL    & \third{20.60} & 10.67 & 13.16 & 5.88 & 9.99 & \third{29.71} & \first{6.43}  & \first{30.85}  & \first{17.51}  & \first{56.72}  & \first{16.30}  & \first{25.82}  \\
\bottomrule
\end{tabular}}

\end{table}

\subsection{Symmetric Real-World Datasets for Link Prediction}
We selected twelve datasets for link prediction, including well-known citation networks such as \textsc{Cora} and \textsc{Citeseer}, large-scale benchmarks from the Open Graph Benchmark suite like \textsc{ogbl-citation2} and \textsc{ogbl-collab}, and structurally complex knowledge and ontology graphs such as \textsc{AIFB}, \textsc{Edit-TSW}. More details about datasets can be found in Appendix \ref{app:dataset-stats}. For each dataset, we compute the proposed symmetry metric $\hat{r}_G$ (Equation \ref{eq:symm_new}). The computed values are reported in the first row of Table~\ref{tab:symm}, with datasets sorted in ascending order of $\hat{r}_G$. As shown, the datasets exhibit widely varying degrees of symmetry. For instance, \textsc{ogbl-citation2}, \textsc{Cora}, and \textsc{Pubmed} display very low $\hat{r}_G$ values, largely due to the presence of informative node features that yield many unique WL colors. At the other end of the spectrum, datasets such as \textsc{Edit-TSW}, which lacks node features and exhibit high topological regularity, show significantly higher symmetry scores.

Table~\ref{tab:symm} reports the results, in terms of the standard link prediction metric MRR, achieved by the models described in Section~\ref{subsec:methods}. The models are listed in ascending order of theoretical expressiveness, as characterized by Theorem~\ref{theo:model}. Since we are now evaluating under the standard link prediction setting (i.e., graphs with missing edges), we also include the NCNC variant of NCN. For ELPH, we report results for BUDDY, its more scalable version. All evaluations are conducted under the challenging and realistic HeaRT setting~\citep{li2023evaluating}. Further details on model architectures and hyperparameter selection are provided in Appendix~\ref{app:experimental-details}, while the complete results, including standard deviations and an additional evaluation metric, are reported in Appendix~\ref{app:results-std}.

The results reveal a clear trend: as $\hat{r}_G$ increases, more expressive models become necessary. In the five datasets with the lowest symmetry, pure GNN-based methods consistently rank among the top three performers. However, as symmetry increases, their performance deteriorates and none of them appears in the top three on the six most symmetric datasets. Conversely, SEAL, the most expressive model in our hierarchy, ranks first across all high-symmetry datasets.
This provides a key practical insight: while simple, less expressive models may suffice when node features are informative and structural ambiguity is limited, datasets with high link symmetry demand more powerful architectures. %High expressiveness is thus not universally beneficial, but becomes crucial in scenarios where structural symmetries hinder the effectiveness of message passing.

\section{Conclusion} \label{conc}
%We presented a principled study of the expressive power of message-passing models for link representation.
We propose a theoretical framework that formally compares a wide range of existing models and establishes a hierarchy of them. We introduced \SyntData, the first synthetic benchmark explicitly designed to evaluate link-level expressiveness. Finally, we demonstrated that increased expressiveness translates into performance gains on real-world link prediction tasks, especially in structurally symmetric graphs. These findings highlight that model selection should account for dataset symmetry: while simple models may suffice in low-symmetry settings, high-expressiveness architectures are essential when structural ambiguity is present.
\paragraph{Limitations and Future Work}
Our framework is tailored to MP models and does not directly account for recent advances in transformer-based architectures or spectral methods for link representation.
Extending the framework to encompass these directions may provide additional insights, making this an interesting and valuable direction for future work. Furthermore, the high-symmetry datasets used in our evaluation are relatively small, highlighting the need for larger and more robust benchmarks to properly assess model generalization and advance research in this area~\cite{bechler2025position}.

\newpage
% \section{todos}
% \AL{todo: Fix the template}\\
% \AL{todo: Add citations}\\
% \AL{Questione Elph e Buddy in fiugra e Tabella}\\
% \AL{It is relay important to specifically assess that the new dataset are small!!!! we must think to a good motivation for this! Of courese as soon as the model complexity increases (i.e. slr models) leads oom in big dataset. Another motivation coulb de that such a graph do not exists in reality?  }
% \VL{provare GAE nel dataset sintetico}
% \VL{ncnc dovrebbe fare di più}
% \VL{non si riesce proprio a migliorare peg?}
% \VL{ma abbiamo usato numero di layer diversi in \SyntData?}
% \VL{aggiungere nella caption delle tabelle commenti dei results}\VL{scrivere i nomi dei modelli uguali dappertutto}
% \VL{avere gli stessi modelli in tutte le tabelle}
% \VL{controlla bene l'uso della parole link o pairs of node. secondo me meglio se metti link ma spieghi all'inizio che con la parola link intendi $\{u,v\}$, non in E ma in VxV}\\
% \VL{DEVI FARE CAPIRE BENE E TROVARE UN MODO GIUSTO PER SCRIVERE LINK ISOMORFI COMPOSTI DA NODI ISOMORFI  }

% \VL{cambiare nel 4 wl con autormorfi}
% \VL{scrivere nell'ultima sezione che includiamo anche le modifiche dei modelli ncnc perchè siamo in link prediction e dì che l'espressività è come quella di ncn}

%\VL{dovrai dire qualcosa sul numero di layer}

\clearpage

\bibliographystyle{plainnat}
\bibliography{bib}

%%%%%%%%%%%%%%%%%%%%%%%%%%%%%%%%%%%%%%%%%%%%%%%%%%%%%%%%%%%%

\clearpage

\newpage
\appendix

\section{Proofs}\label{app:proofs}
\begin{proof}[\textbf{Proof of Theorem \ref{theo:fram}}] We prove theorem by addressing each of its three components individually.

\begin{enumerate}

\item Let $M\in \mathcal{M}$. If $m^M=0$, then, regardless of $k_\phi^M$, $M$ is not able to distinguish between links whose endpoints are automorphic, i.e., 
\begin{align}
\forall F\in M, \forall (u, v), (u', v') \text{ s.t. } \exists \sigma_1, \sigma_2 &\in \Sigma_n^G \text{ with } \sigma_1(u) = u' \land \sigma_2(v) = v'\\
F((u', v'), \mathbf{X}^0) &= F((u, v),\mathbf{X}^0)
\end{align}
therefore, automorphic links whose endpoints are automorphic will be assigned to the same representation. Moreover, given $M_1,M_2 \in \mathcal{M}$, if $m^{M_1}=0$, while $m^{M_2}>0$ and $\text{\small COMB}$ is injective, regardless of $k^{M_1}_{\phi}$ and $k^{M_2}_{\phi}$ , $M_1 \leq M_2$.

\textit{\textbf{Proof}} Suppose $m^M=0$, i.e., $M$ is composed only of the functions $\phi$ and $g$, i.e., its functions are:
\[
F((u, v), \mathbf{X}^0) = g\big(\phi(u, \mathbf{X}^0), \phi(v, \mathbf{X}^0)\big).
\]
Assume $\phi$ is maximally expressive at the node level (note that no GNN can achieve this, as the $k$-WL algorithm, preserves graph automorphisms for every $k$ \citep{lichter2025computational,dawar2020generalizations,cai1992optimal}), i.e., it assigns representations to nodes solely based on their automorphism classes. Then for any automorphisms $\sigma_1, \sigma_2 \in \Sigma_n^G$ such that $\sigma_1(u) = u'$ and $\sigma_2(v) = v'$, it follows that
\[
\phi(u, \mathbf{X}^0) = \phi(u', \mathbf{X}^0), \quad \phi(v, \mathbf{X}^0) = \phi(v',\mathbf{X}^0).
\]
Thus, regardless of the function $g$, we obtain
\[
g\big(\phi(u,  \mathbf{X}^0), \phi(v, \mathbf{X}^0)\big) = g\big(\phi(u', \mathbf{X}^0), \phi(v', \mathbf{X}^0)\big),
\]
i.e., the model cannot distinguish between links whose endpoints are automorphic. If $(u,v)$ and $(u',v')$ are not themselves automorphic pairs, then the model fails to differentiate them, thus assigning the same representation to non-automorphic links.

Now consider $M_1 \in \mathcal{M}$ with all functions $F\in M_1$ composed only of $\phi$ and $g$, while $M_2 \in \mathcal{M}$ with all functions $F\in M_2$ that also include a component $f$ (on the $m$-order neighborhood). Using Definition \ref{def:moreex}, we show that $M_2$ is more expressive.

Let $F_1\in M_1$, $F_2\in M_2$, $(u,v), (u',v') \in V_G \times V_G$ such that $(u,v) \not \simeq(u',v')$ and
\[
F_1((u,v), \mathbf{X}^0) = F_1((u',v'), \mathbf{X}^0),
\]
but
\[
F_2((u,v), \mathbf{X}^0) \neq F_2((u',v'), \mathbf{X}^0).
\]
This implies that
\[
g\big(\phi(u, \mathbf{X}^0), \phi(v, \mathbf{X}^0)\big) = g\big(\phi(u', \mathbf{X}^0), \phi(v', \mathbf{X}^0)\big),
\]
but due to the presence of a component $f$ and the use of an injective COMB, we get
\[
F_2((u,v), \mathbf{X}^0) \neq F_2((u',v'), \mathbf{X}^0),
\]
demonstrating by Definition \ref{def:moreex} that $M_2$ is more expressive than $M_1$.

\item Let $M_1,M_2 \in \mathcal{M}$ with $k^{M_1}_{{\rho}}=k^{M_2}_{{\rho}}=\textrm{1-WL}$ and let $l^{M_1},l^{M_2}$ be the number of layers used respectively by $M_1,M_2$. If $m^{M_1}+l^{M_1} \leq m^{M_2}+l^{M_2}$, then $M_1\leq M_2$.

\textit{\textbf{Proof}} Let $F_1\in M_1$, $F_2\in M_2$. Assume $l^{M_2} \geq l^{M_1} + m^{M_1} - m^{M_2}$ and $l^{M_1} \leq l^{M_2}$ and $m^{M_1} > m^{M_2}$.

Let $(u,v), (u',v') \in V_G \times V_G$ such that
\[
F_1((u,v), G, \mathbf{X}^0) \neq F_1((u',v'), G, \mathbf{X}^0).
\]
Let $c_u^\ell$ denote the color assigned to node $u$ after $\ell$ iterations of the 1-WL algorithm. Assume $c_u^\ell = c_{u'}^\ell$ and $c_v^\ell = c_{v'}^\ell$; otherwise, $F_2$ alone could distinguish between $(u,v)$ and $(u',v')$, and the proof would be complete.

We divide the proof into three steps. Since $F_1$ distinguishes $(u,v)$ from $(u',v')$ while $c_u^\ell = c_{u'}^\ell$ and $c_v^\ell = c_{v'}^\ell$, this implies that:
\[
\left\{ \rho_1(w, \mathbf{X}) \,\middle|\, w \in \bigcup_{j=0}^{m^{M_1}} N^j(u,v) \right\}
\neq
\left\{ \rho_1(i, \mathbf{X}) \,\middle|\, i \in \bigcup_{j=0}^{m^{M_1}} N^j(u',v') \right\} \tag{1}
\]

Let's consider the smaller neighborhood $m^{M_2}$:
\[
\left\{ \rho_1(w, \mathbf{X}) \,\middle|\, w \in \bigcup_{j=0}^{m^{M_2}} N^j(u,v) \right\}
\quad \text{and} \quad
\left\{ \rho_1(i, \mathbf{X}) \,\middle|\, i \in \bigcup_{j=0}^{m^{M_2}} N^j(u',v') \right\}.
\]

If these sets are already different, then $F_2$ (with smaller neighborhood but same $\rho_1$) suffices, and the proof is complete. If instead they are equal, then from (1) it follows that the difference lies in the sets:
\[
\left\{ \rho_1(w, \mathbf{X}) \,\middle|\, w \in \bigcup_{j=m_2+1}^{m^{M_1}} N^j(u,v) \right\}
\neq
\left\{ \rho_1(i, \mathbf{X}) \,\middle|\, i \in \bigcup_{j=m_2+1}^{m^{M_1}} N^j(u',v') \right\}.
\]

Since $F_2$ uses a number of layers $l^{M_2} \geq l^{M_1} + m^{M_1} - m^{M_2}$, it can compensate for the smaller neighborhood radius $m_2$ by using more layers to reach the same depth of information as $F_1$. Thus, using $\rho_2$ in $F_2$, we obtain:
\[
\left\{ \rho_2(w, \mathbf{X}) \,\middle|\, w \in \bigcup_{j=0}^{m^{M_2}} N^j(u,v) \right\}
\neq
\left\{ \rho_2(w, \mathbf{X}) \,\middle|\, w \in \bigcup_{j=0}^{m^{M_2}} N^j(u',v') \right\},
\]
which implies
\[
F_2((u,v), \mathbf{X}^0) \neq F_2((u',v'), \mathbf{X}^0),
\]
and by Definition \ref{def:moreex}, this yields $M_1 \leq M_2$.

\item Let $M_1,M_2 \in \mathcal{M}$ with $m^{M_1} = m^{M_2}$. If $k^{M_1}_{\rho} \leq k^{M_2}_{\rho}$, then $M_1 \leq M_2$.

\textit{\textbf{Proof}} Let $m^{M_1} = m^{M_2}=m$ and $k^{M_1}_{\rho} \leq k^{M_2}_{\rho}$. Let $F_1\in M_1$ and $ F_2\in M_2$ and $(u,v), (u',v') \in V_G \times V_G$ such that
\[
F_1((u,v), G, \mathbf{X}^0) \neq F_1((u',v'), G, \mathbf{X}^0).
\]

Assume that $c_u^\ell = c_{u'}^\ell$ and $c_v^\ell = c_{v'}^\ell$, i.e., $u$ and $u'$, as well as $v$ and $v'$, have the same WL color after $\ell$ iterations. Otherwise, $F_2$ could already distinguish the links based on $k^{M_2}_{\rho}$ alone, and the proof would be trivial.

Now, since $F_1$ can distinguish the links despite $c_u^\ell = c_{u'}^\ell$ and $c_v^\ell = c_{v'}^\ell$, this implies that:
\[
\left\{ \rho_1(w, G, \mathbf{X}) \,\middle|\, w \in \bigcup_{j=0}^{m} N^j(u,v) \right\}
\neq
\left\{ \rho_1(i, G, \mathbf{X}) \,\middle|\, i \in \bigcup_{j=0}^{m} N^j(u',v') \right\} \tag{1}
\]

Since $k^{M_1}_{\rho} \leq k^{M_2}_{\rho}$, and both models operate on the same neighborhood radius $m$, it follows that there exists a number of layers (possibly greater than in $F_1$ \citep{grohe2025iteration,immerman2019k}) such that $F_2$ can also distinguish these sets. That is,
\[
\left\{ \rho_2(w, G, \mathbf{X}) \,\middle|\, w \in \bigcup_{j=0}^{m} N^j(u,v) \right\}
\neq
\left\{ \rho_2(i, G, \mathbf{X}) \,\middle|\, i \in \bigcup_{j=0}^{m} N^j(u',v') \right\},
\]
which implies
\[
F_2((u,v), G, \mathbf{X}^0) \neq F_2((u',v'), G, \mathbf{X}^0).
\]

Therefore, by Definition \ref{def:moreex}, $F_1 \leq F_2$.

\end{enumerate}

\end{proof}

\begin{proof}[\textbf{Proof of Theorem \ref{theo:model}}]
\textbf{1.} This follows directly from Theorem \ref{theo:fram}. Pure GNNs only rely on node representations $\phi$ and the combining function $g$, without any function $f$ over the $m$-order neighborhood. As established in Theorem \ref{theo:fram}, such models cannot distinguish between links whose endpoints are automorphic. In contrast, NCN, Neo-GNN, ELPH, and SEAL incorporate neighborhood-level functions $f$ (all with injective $\text{\small COMB}$ operations, namely concatenation), which increases their expressive power.

\textbf{2.} According to Table \ref{tab:fram}, Neo-GNN uses a fixed number of message-passing layers $l^{\text{Neo-GNN}} = 1$, while NCN allows $l^{\text{NCN}}$ to be freely chosen. Since both models operate with the same type of base function $\rho$ (e.g., 1-WL), Theorem \ref{theo:fram} implies that if $l^{\text{NCN}} \geq m^{\text{Neo-GNN}}$, then NCN can simulate or surpass Neo-GNN’s receptive field and representational capability. Hence, $\text{Neo-GNN} \leq \text{NCN}$.

\textbf{3.} By design, ELPH does not use any message-passing layers ($l^{\text{ELPH}} = 0$, see Table \ref{tab:fram}), and instead aggregates counts over the $m$-hop neighborhood. If Neo-GNN uses a neighborhood of size at least $m^{\text{ELPH}} -1$, then Theorem \ref{theo:fram} ensures it can access all the information available to ELPH and apply additional learned transformations via its GNN layers. Thus, Neo-GNN is more expressive.

\textbf{4.} ELPH relies only on aggregated neighborhood counts and uses $l^{\text{ELPH}} = 0$ layers. NCN uses a fixed radius $m^{\text{NCN}} = 1$ (Table \ref{tab:fram}), but allows a configurable number of layers. If $l^{\text{NCN}} \geq m^{\text{ELPH}} - 1$, then by Theorem \ref{theo:fram} NCN can effectively capture structural signals that ELPH computes with wider but shallower architectures, achieving greater expressiveness.

\textbf{5.} SEAL is more expressive due to both its flexible architecture and enhanced message-passing capabilities. Specifically, SEAL builds a subgraph around the link and augments the adjacency matrix with positional encodings that are link-aware, effectively allowing a learned structural encoding over the link neighborhood $N^m(u,v)$. This corresponds to the expressive power of a higher-order WL test (namely $1$-$|N^m(u,v)|$-WL \citep{zhou2023relational}), which is strictly more powerful than 1-WL. Moreover, SEAL allows arbitrary $m^{\text{SEAL}}$ and $l^{\text{SEAL}}$, so for any other model with weaker $\rho$, one can always configure SEAL with the same $m$ and a sufficient number of layers to match or exceed its expressiveness. Prior results have shown that highly expressive architectures may require deeper message passing to converge \citep{grohe2025iteration,immerman2019k}, which SEAL supports by design.
\end{proof}

\section{Additional Information on \SyntData}

\subsection{Evaluation method}
\label{app:synthetic_setting}

This section provides a more detailed explanation of how the two components of the Reliable Paired Comparison work to evaluate different models in the synthetic setting.

\paragraph{Major Procedure}
Given a graph $G$, we select two structurally distinct links $e=(u,v), e'=(u',v')$.
Then we create $q$ isomorphic copies $\{G_1, G_2, \dots, G_q\}$ by randomly permuting its nodes. For each copy $G_i$ and pair of links $e_i, e'_i$, we compute the embedding differences:
\begin{equation}
    d_i = f(G_i, e_i) - f(G_i, e'_i), \quad i \in [q],
\end{equation}

where $f$ is the embedding function learned by the model. Assuming that the difference vectors are independent random vectors $\mathcal{N}(\mu, \Sigma)$, we consider that $f(G_i)$
follows a Gaussian distribution, so random permutations only introduce Gaussian noise into the results. %\MJ{I am not sure I understand the role of the permutations here}

If the model is not able to distinguish links $e$ and $e'$, the mean difference should be $\mu = 0$. To check this a $\alpha-$level Hotelling's T-square test is conducted, comparing hypothesis $H_0 : \mu = 0$ and $H_1 : \mu \not= 0$. The $T^2-$statistic for $\mu$ is calculed as:
\begin{equation}
T^2 = q (\bar{d} - \mu)^T S^{-1} (\bar{d} - \mu),
\label{eq:t2}
\end{equation}

where

\begin{equation}
    \bar{d} = \frac{1}{q} \sum_{i=1}^{q} d_i, \quad 
    \mathbf{S} = \frac{1}{q - 1} \sum_{i=1}^{q} (d_i - \bar{d}) (d_i - \bar{d})^T.
\end{equation}

Hotelling’s T-square test establishes that $T^2$ follows the distribution of a random variable  
$\frac{(q-1)d}{q-d} F_{d, q-d}$, where $F_{d, q-d}$ represents an $F$-distribution with degrees of freedom $d$ and $q - d$ \cite{hotelling1931generalization}.  
This theorem provides a link between the unknown parameter $\mu$ and a specific distribution $F_{d, q-d}$,  
enabling us to validate the confidence interval of $\mu$ by analyzing how well it matches the assumed distribution.  
To evaluate the hypothesis $H_0: \boldsymbol{\mu} = 0$, we substitute $\boldsymbol{\mu} = 0$ into Equation (\ref{eq:t2}), leading to:

\begin{equation}
    T^2_{\text{test}} = q \bar{\boldsymbol{d}}^T S^{-1} \bar{\boldsymbol{d}}.
\end{equation}

An $\alpha$-level test of $H_0: \boldsymbol{\mu} = 0$ against the alternative $H_1: \boldsymbol{\mu} \neq 0$  
supports $H_0$ (meaning the model fails to differentiate the pair) if:

\begin{equation}
    T^2_{\text{test}} = q \bar{\boldsymbol{d}}^T S^{-1} \bar{\boldsymbol{d}} < \frac{(q - 1) d}{(q - d)} F_{d, q - d}(\alpha),
\end{equation}

where $F_{d, q - d}(\alpha)$ represents the upper $(100\alpha)$-th percentile of the $F$-distribution $F_{d, q - d}$  
\cite{Fisher1950}, with degrees of freedom $d$ and $q - d$.  
Conversely, we reject $H_0$ (indicating that the model successfully distinguishes the pair) if:

\begin{equation}
    T^2_{\text{test}} = q \bar{\boldsymbol{d}}^T S^{-1} \bar{\boldsymbol{d}} > \frac{(q - 1) d}{(q - d)} F_{d, q - d}(\alpha).
\end{equation}

\paragraph{Reliability Check} By selecting an appropriate value of $\alpha$, the Main Procedure ensures a reliable confidence interval to evaluate distinguishability. However, choosing $\alpha$ heuristically may not be the optimal approach. Additionally, computational precision can introduce deviations in the assumed Gaussian fluctuations. To address this issue, we employ the Reliability Check, which captures both external differences between two graphs and internal variations within a single graph.

Without loss of generality (WLOG), we replace $(G_i, e'_i)$ with a permutation of $G$, denoted as $(G^\pi, e^\pi)$. This allows us to analyze the internal fluctuations of $G$ for the same link. We follow the same steps as in the Main Procedure.

Proceeding with $(G_i, e_i)$ and $(G_i^\pi, e_i^\pi)$, we compute the $T^2$-statistic as:

\begin{equation}
T^2_{\text{reliability}} = q \bar{d}^T S^{-1} \bar{d},
\end{equation}

where

\begin{equation}
\bar{d} = \frac{1}{q} \sum_{i=1}^{q} d_i, \quad d_i = f(G_i, e_i) - f(G_i^\pi, e_i^\pi), \quad i \in [q],
\end{equation}

\begin{equation}
S = \frac{1}{q - 1} \sum_{i=1}^{q} (d_i - \bar{d}) (d_i - \bar{d})^T.
\end{equation}

Since $G$ and $G^\pi$ are isomorphic, the GNN should not differentiate between them, implying that $\mu = 0$. Consequently, the test is considered reliable only if

\begin{equation}
T^2_{\text{reliability}} < \frac{(q-1)d}{(q-d)} F_{d, q-d}(\alpha).
\end{equation}

By combining the reliability check with the distinguishability results, we obtain the full RPC Reliability Procedure Check.

For each pair of structurally different links $e$ and $e'$, the threshold is computed as: 

\begin{equation}
\text{Threshold} = \frac{(q-1)d}{(q-d)} F_{d,q-d}(\alpha).
\end{equation}

Next, we perform the Main Procedure on  $e$ and $e'$ to evaluate distinguishability, and we conduct the Reliability Check on $e$ and $e^\pi$. The GNN is considered capable of distinguishing $e$ and $e'$ only if both conditions

\begin{equation}
T^2_{\text{reliability}} < \text{Threshold} < T^2_{\text{test}}
\end{equation}

are satisfied.

\section{Experimental details}\label{app:experimental-details}

\subsection{Real world datasets}\label{app:dataset-stats}

Table~\ref{tab:dataset-stats-standard} presents the statistics of the standard datasets commonly used for link prediction tasks. The Cora, Citeseer, and Pubmed datasets are well-known citation networks, often employed as benchmarks for graph-based learning methods. These datasets are relatively small in scale, both in terms of nodes and edges. In contrast, the OGB (Open Graph Benchmark) datasets including ogbl-collab, ogbl-ddi, ogbl-ppa, and ogbl-citation2 are significantly larger and more complex, providing challenging benchmarks for evaluating scalability and performance on large graphs. 

For Cora, Citeseer, and Pubmed, we adopt a fixed train/validation/test split of 85/5/10\%. For the OGB datasets, we use the official splits provided by the OGB benchmark.

\begin{table}[ht]
\setlength{\tabcolsep}{4pt}
\footnotesize
\centering
\begin{tabular}{lccccccc}
\toprule
\textbf{Dataset} & Cora & Citeseer & Pubmed & ogbl-collab & ogbl-ddi & ogbl-ppa & ogbl-citation2 \\
\midrule
\#Nodes & 2,708 & 3,327 & 18,717 & 235,868 & 4,267 & 576,289 & 2,927,963 \\
\#Edges & 5,278 & 4,676 & 44,327 & 1,285,465 & 1,334,889 & 30,326,273 & 30,561,187 \\
Mean Degree & 3.90 & 2.81 & 4.74 & 10.90 & 625.68 & 105.25 & 20.88 \\
Split Ratio & 85/5/10 & 85/5/10 & 85/5/10 & 92/4/4 & 80/10/10 & 70/20/10 & 98/1/1 \\
\bottomrule
\end{tabular}
\caption{Statistics of datasets. The split ratio is for train/validation/test.}
\label{tab:dataset-stats-standard}
\end{table}

Table~\ref{tab:dataset-stats-symmetric} reports the statistics of additional datasets characterized by high symmetry, which are also used in our experiments. These datasets span various domains, including social networks, biological systems, and knowledge graphs, offering diverse structural properties and serving as valuable benchmarks for evaluating models under highly regular and repetitive connection patterns.

Specifically, the \textit{AIFB}\footnote{\url{https://pytorch-geometric.readthedocs.io/}} dataset models the organizational structure of the AIFB research institute, capturing relationships among its staff, research groups, and publications. %\textit{Mutag}\footnotemark[1] is a molecular graph dataset widely used for chemical compound analysis. %\textit{Maayan-Vidal}\footnote{\url{http://konect.cc/}} represents a metabolic network, providing insights into biological processes. 
\textit{Edit-TSW}\footnotemark[2] models user interactions and content editing activities within the Wiktionary platform. 
Finally, the \textit{NSC}\footnote{\url{https://github.com/LeiCaiwsu/LGLP/tree/main/LGLP/Python/data}}, 
\textit{YST}\footnotemark[3], and \textit{GRQ}\footnotemark[3] datasets, introduced in~\cite{cai2021line}, focus on relational learning tasks and exhibit highly symmetric structures, making them particularly suitable for evaluating link prediction models.

\begin{table}[ht]
\setlength{\tabcolsep}{4pt}
\footnotesize
\centering
\begin{tabular}{lccccc}
\toprule
\textbf{Dataset} & NSC & YST & GRQ  & aifb & edit-tsw\\
\midrule
\#Nodes          & 332 & 2,284 & 5,241  & 8,285 & 1,079 \\
\#Edges          & 2,126 & 6,646 & 14,484&  46,042& 2.756  \\
Mean Degree      & 12.81 & 5.82 & 5.53&  5.56& 5.11\\
Split Ratio      & 80/10/10& 80/10/10& 80/10/10& 80/10/10& 80/10/10  \\
\bottomrule
\end{tabular}
\caption{Statistics of datasets. The split ratio is for train/validation/test.}
\label{tab:dataset-stats-symmetric}
\end{table}

\subsection{Implementation details}\label{subsec:implementation-details}
This section provides hyperparameters details for all models trained and evaluated on both \SyntData and the real-world datasets.
All experiments were conducted on a workstation running Ubuntu 22.04 with an AMD Ryzen 9 7950X CPU (32 threads), 124GB of RAM, and two NVIDIA GeForce RTX 4090 GPUs (24GB each).

\paragraph{Synthetic Hyperparameter Settings}

 We present the hyperparameter search space for the \SyntData datasets in Table \ref{tab:hyperparam-syntetic}. For each model, we initially followed the hyperparameter configurations proposed in their respective papers. However, due to the small size of the synthetic graphs in \SyntData, we significantly reduced the embedding dimensions, which led to improved performance in most cases.

\begin{table}[ht]
\centering
\setlength{\tabcolsep}{4pt}
\footnotesize
\begin{tabular}{lcccccc}
\toprule
\textbf{Dataset} & \textbf{LR} & \textbf{Drop.} & \textbf{WD} & \#\textbf{L} & \#\textbf{P} & \textbf{Dim} \\
\midrule
\SyntData & (0.01, 0.001) & (0.1, 0.3, 0.5) & (1e-4, 1e-7, 0) & (1, 2, 3) & (1, 2, 3) & (2–256) \\
\bottomrule
\end{tabular}
\caption{Hyperparameter Search Ranges for \SyntData. 
Abbreviations: LR = Learning Rate, Drop. = Dropout, WD = Weight Decay, \#L = Number of Model Layers, \#P = Number of Prediction Layers, Dim = Embedding Dimension.}
\label{tab:hyperparam-syntetic}
\end{table}

\paragraph{Real-World Hyperparameter Settings}
We report the hyperparameter search space for all real-world datasets used in our experiments in Table~\ref{tab:hyperparam-real}. 
For \textit{Cora}, \textit{Citeseer}, \textit{Pubmed}, \textit{ogbl-collab}, \textit{ogbl-ddi}, \textit{ogbl-ppa}, and \textit{ogbl-citation2}, we adopt the hyperparameter settings declared in the  paper~\citep{li2023evaluating}. 
For the remaining datasets, \textit{NSC}, \textit{YST}, \textit{GRQ}, \textit{aifb}, and \textit{edit-tsw}, we conducted a hyperparameter search using the ranges specified in the table.

\begin{table}[ht]
\centering
\setlength{\tabcolsep}{4pt}
\footnotesize
\begin{tabular}{lcccccc}
\toprule
\textbf{Dataset} & \textbf{LR} & \textbf{Drop.} & \textbf{WD} & \#\textbf{L} & \#\textbf{P} & \textbf{Dim} \\
\midrule
Cora            & (0.01, 0.001)     & (0.1, 0.3, 0.5)     & (1e-4, 1e-7, 0)   & (1, 2, 3)   & (1, 2, 3)   & (128, 256) \\
Citeseer        & (0.01, 0.001)     & (0.1, 0.3, 0.5)     & (1e-4, 1e-7, 0)   & (1, 2, 3)   & (1, 2, 3)   & (128, 256) \\
Pubmed          & (0.01, 0.001)     & (0.1, 0.3, 0.5)     & (1e-4, 1e-7, 0)   & (1, 2, 3)   & (1, 2, 3)   & (128, 256) \\
ogbl-collab     & (0.01, 0.001)    & (0, 0.3, 0.5)     & 0      & 3   & 3   & 256 \\
ogbl-ddi        & (0.01, 0.001)    & (0, 0.3, 0.5)     & 0      & 3   & 3   & 256 \\
ogbl-ppa        & (0.01, 0.001)    & (0, 0.3, 0.5)     & 0      & 3   & 3   & 256 \\
ogbl-citation2  & (0.01, 0.001)    & (0, 0.3, 0.5)     & 0      & 3   & 3   & 128 \\
NSC            & (0.01, 0.001) & (0.1, 0.3, 0.5) & (1e-4, 1e-7, 0) & (1, 2, 3) & (1, 2, 3) & (64, 128, 256) \\
YST            & (0.01, 0.001) & (0.1, 0.3, 0.5) & (1e-4, 1e-7, 0) & (1, 2, 3) & (1, 2, 3) & (64, 128, 256) \\
GRQ            & (0.01, 0.001) & (0.1, 0.3, 0.5) & (1e-4, 1e-7, 0) & (1, 2, 3) & (1, 2, 3) & (64, 128, 256) \\
aifb           & (0.01, 0.001) & (0.1, 0.3, 0.5) & (1e-4, 1e-7, 0) & (1, 2, 3) & (1, 2, 3) & (64, 128, 256) \\
edit-tsw       & (0.01, 0.001) & (0.1, 0.3, 0.5) & (1e-4, 1e-7, 0) & (1, 2, 3) & (1, 2, 3) & (64, 128, 256) \\
\bottomrule
\end{tabular}
\caption{
Hyperparameter search ranges for all real-world datasets. Abbreviations: LR = Learning Rate, Drop. = Dropout, WD = Weight Decay, \#L = Number of Model Layers, \#P = Number of Prediction Layers, Dim = Embedding Dimension.
}
\label{tab:hyperparam-real}
\end{table}

\section{Real-world results with standard deviations}\label{app:results-std}
In this section, we present the same results as in Table~\ref{tab:symm}, now including standard deviations computed over 5 runs with different random seeds. Specifically, Table~\ref{tab:mrr-part1} reports the MRR results on standard small datasets, Table~\ref{tab:mrr-part2} shows the MRR results on the OGB benchmark, and Table~\ref{tab:mrr-part3} reports the MRR results for highly symmetric datasets.

\begin{table}[ht]
\caption{MRR with standard deviations on real-world link prediction datasets.}
\label{tab:mrr-part1}
\centering
\setlength{\tabcolsep}{4pt}
\footnotesize
\begin{tabular}{lccccc}
\toprule
\textbf{Models} & \textsc{NSC} & \textsc{YST} & \textsc{GRQ} & \textsc{aifb} & \textsc{edit-tsw} \\
$\hat{r}_G$ & 0.16 & 0.24 & 0.30 & 0.41 & 0.67 \\
\midrule
\GCN    & \meanstd{26.79}{1.08} & \meanstd{1.17}{0.02} & \meanstd{6.89}{0.47} & \meanstd{10.85}{0.12} & \meanstd{4.12}{0.21} \\
\GAT    & \meanstd{2.47}{3.58}  & \meanstd{0.40}{0.00} & \meanstd{0.63}{0.40} & \meanstd{1.07}{0.51}  & \meanstd{0.40}{0.00} \\
\SAGE   & \meanstd{20.32}{0.11} & \meanstd{1.77}{1.38} & \meanstd{1.34}{1.32} & \meanstd{0.54}{0.08}  & \meanstd{8.35}{7.27} \\
\GAE    & \meanstd{16.61}{10.00} & \meanstd{2.32}{0.01} & \meanstd{2.32}{0.01} & \meanstd{10.98}{0.01} & \meanstd{3.54}{0.01} \\
\BUDDY  & \meanstd{18.40}{0.07} & \second{\meanstd{13.68}{0.61}} & \meanstd{46.23}{0.42} & \second{\meanstd{13.73}{0.11}} & \second{\meanstd{22.66}{0.46}} \\
\ELPH   & \meanstd{26.09}{1.04} & \meanstd{13.15}{1.12} & \meanstd{38.88}{0.55} & \third{\meanstd{13.39}{0.60}} & \meanstd{10.10}{7.37} \\
\NEOGNN & \meanstd{22.86}{0.75} & \meanstd{5.78}{0.12}  & \meanstd{29.51}{0.38} & \meanstd{13.11}{0.51} & \meanstd{8.10}{0.29} \\
\NCN    & \second{\meanstd{30.36}{0.28}} & \meanstd{11.99}{0.06} & \second{\meanstd{48.45}{0.02}} & \meanstd{6.47}{0.23}  & \third{\meanstd{10.32}{0.39}} \\
\NCNC   & \third{\meanstd{30.20}{0.03}}  & \third{\meanstd{13.54}{0.47}} & \third{\meanstd{47.60}{0.52}} & \texttt{OOM} & \meanstd{10.30}{0.39} \\
\SEAL   & \first{\meanstd{30.85}{1.46}} & \first{\meanstd{17.51}{0.94}} & \first{\meanstd{56.72}{1.35}} & \first{\meanstd{16.30}{0.22}} & \first{\meanstd{25.82}{1.47}} \\
\bottomrule
\end{tabular}
\end{table}

\begin{table}[ht]
\caption{MRR with standard deviations on real-world link prediction datasets.}
\label{tab:mrr-part2}
\centering
\setlength{\tabcolsep}{4pt}
\footnotesize
\begin{tabular}{lccc}
\toprule
\textbf{Models} & \textsc{Cora} & \textsc{Citeseer} & \textsc{Pubmed} \\
$\hat{r}_G$ & 0.02 & 0.02 & 0.08 \\
\midrule
\GCN     & \second{\meanstd{16.61}{0.30}} & \meanstd{21.09}{0.88} & \meanstd{7.13}{0.27} \\
\GAT     & \meanstd{13.84}{0.68} & \meanstd{19.58}{0.84} & \meanstd{4.95}{0.14} \\
\SAGE    & \meanstd{14.74}{0.69} & \meanstd{21.09}{1.15} & \second{\meanstd{9.40}{0.70}} \\
\GAE     & \first{\meanstd{18.32}{0.41}} & \first{\meanstd{25.25}{0.82}} & \meanstd{5.27}{0.25} \\
\BUDDY   & \meanstd{13.71}{0.59} & \second{\meanstd{22.84}{0.36}} & \meanstd{7.56}{0.18} \\
\NEOGNN  & \meanstd{13.95}{0.39} & \meanstd{17.34}{0.84} & \meanstd{7.74}{0.30} \\
\NCN     & \meanstd{14.66}{0.95} & \third{\meanstd{28.65}{1.21}} & \meanstd{5.84}{0.22} \\
\NCNC    & \third{\meanstd{14.98}{1.00}} & \meanstd{24.10}{0.65} & \third{\meanstd{8.58}{0.59}} \\
\SEAL    & \meanstd{10.67}{3.46} & \meanstd{13.16}{1.66} & \meanstd{5.88}{0.53} \\
\bottomrule
\end{tabular}
\end{table}

\begin{table}[t]
\caption{MRR with standard deviations on real-world link prediction datasets.}
\label{tab:mrr-part3}
\centering
\setlength{\tabcolsep}{4pt}
\footnotesize
\begin{tabular}{lcccc}
\toprule
\textbf{Models} & \textsc{ogbl-citation2} & \textsc{ogbl-ddi} & \textsc{ogbl-ppa} & \textsc{ogbl-collab} \\
$\hat{r}_G$ & 0.01 & 0.10 & 0.12 & 0.13 \\
\midrule
\GCN     & \meanstd{19.98}{0.35} & \first{\meanstd{13.46}{0.34}} & \meanstd{26.94}{0.48} & \meanstd{6.09}{0.38} \\
\GAT     & \second{\meanstd{22.05}{0.12}} & \second{\meanstd{12.92}{0.39}} & \meanstd{27.27}{0.30} & \meanstd{4.18}{0.33} \\
\SAGE    & \texttt{OOM} & \third{\meanstd{12.60}{0.72}} & \texttt{OOM} & \meanstd{5.53}{0.50} \\
\GAE     & \texttt{OOM} & \meanstd{3.49}{1.73} & \texttt{OOM} & \texttt{OOM} \\
\BUDDY   & \meanstd{19.17}{0.20} & \meanstd{12.43}{0.50} & \meanstd{27.00}{0.33} & \meanstd{5.67}{0.36} \\
\NEOGNN  & \meanstd{16.12}{0.25} & \meanstd{10.86}{2.16} & \meanstd{21.68}{1.14} & \meanstd{5.23}{0.90} \\
\NCN     & \first{\meanstd{23.35}{0.28}} & \meanstd{12.86}{0.78} & \first{\meanstd{35.06}{0.26}} & \meanstd{5.09}{0.38} \\
\NCNC    & \meanstd{19.61}{0.54} & \texttt{>24h} & \second{\meanstd{33.52}{0.26}} & \meanstd{4.73}{0.86} \\
\SEAL    & \third{\meanstd{20.60}{1.28}} & \meanstd{9.99}{0.90} & \third{\meanstd{29.71}{0.71}} & \first{\meanstd{6.43}{0.32}} \\
\bottomrule
\end{tabular}
\end{table}

\paragraph{Hits@k.} We also report the Hits@$k$ metric, using $k = 10$ for the standard small datasets (Table~\ref{tab:hits10-part1}) and for the new highly symmetric dataset (Table~\ref{tab:hits10-part2}), and $k = 20$ for the larger datasets (Table~\ref{tab:hits20-part3}), following the evaluation protocol of \citet{li2023evaluating}.

\begin{table}[t]
\caption{Hits@10 with standard deviations on real-world link prediction datasets.}
\label{tab:hits10-part1}
\centering
\setlength{\tabcolsep}{4pt}
\footnotesize
\begin{tabular}{lccccc}
\toprule
\textbf{Models} & \textsc{NSC} & \textsc{YST} & \textsc{GRQ} & \textsc{aifb} & \textsc{edit-tsw} \\
$\hat{r}_G$ & 0.16 & 0.24 & 0.30 & 0.41 & 0.67 \\
\midrule
\GCN    & \meanstd{43.08}{1.36} & \meanstd{0.75}{0.00} & \meanstd{13.47}{2.97} & \meanstd{17.56}{0.92} & \meanstd{5.91}{0.97} \\
\GAT    & \meanstd{2.36}{4.09}  & \meanstd{0.00}{0.00} & \meanstd{0.44}{0.76}  & \meanstd{1.51}{1.73} & \meanstd{0.00}{0.00} \\
\SAGE   & \meanstd{24.06}{3.68} & \meanstd{3.56}{3.96} & \meanstd{2.44}{4.17}  & \meanstd{0.41}{0.38} & \meanstd{10.67}{10.07} \\
\GAE    & \meanstd{30.38}{9.90} & \meanstd{3.38}{0.00} & \meanstd{3.38}{0.00}  & \meanstd{16.92}{0.09} & \meanstd{3.80}{0.00} \\
\BUDDY  & \meanstd{28.93}{2.68} & \second{\meanstd{22.89}{0.69}} & \meanstd{67.26}{0.25} & \meanstd{24.28}{0.57} & \meanstd{36.00}{2.27} \\
\ELPH   & \meanstd{38.37}{0.72} & \meanstd{22.34}{1.15} & \meanstd{59.44}{0.99}  & \second{\meanstd{26.19}{1.83}} & \meanstd{21.45}{15.19} \\
\NEOGNN & \meanstd{37.11}{0.72} & \meanstd{10.29}{0.17} & \meanstd{44.50}{0.24}  & \third{\meanstd{24.29}{1.74}} & \meanstd{12.36}{1.09} \\
\NCN    & \second{\meanstd{47.17}{0.94}} & \meanstd{20.03}{0.00} & \second{\meanstd{68.67}{0.14}} & \meanstd{9.42}{0.76} & \meanstd{16.48}{0.56} \\
\NCNC   & \third{\meanstd{46.86}{0.27}}  & \third{\meanstd{22.44}{1.04}} & \third{\meanstd{68.28}{0.23}} & \texttt{OOM} & \meanstd{15.03}{1.87} \\
\SEAL   & \first{\meanstd{51.26}{0.98}} & \first{\meanstd{28.36}{1.46}} & \first{\meanstd{71.64}{1.04}} & \first{\meanstd{31.84}{0.70}} & \first{\meanstd{42.55}{2.39}} \\
\bottomrule
\end{tabular}
\end{table}

\begin{table}[t]
\caption{Hits@10 with standard deviations on real-world link prediction datasets.}
\label{tab:hits10-part2}
\centering
\setlength{\tabcolsep}{4pt}
\footnotesize
\begin{tabular}{lccc}
\toprule
\textbf{Models} & \textsc{Cora} & \textsc{Citeseer} & \textsc{Pubmed} \\
$\hat{r}_G$ & 0.02 & 0.02 & 0.08 \\
\midrule
\GCN     & \second{\meanstd{36.26}{1.14}} & \meanstd{47.23}{1.88} & \meanstd{15.22}{0.57} \\
\GAT     & \meanstd{32.89}{1.27} & \meanstd{45.30}{1.30} & \meanstd{9.99}{0.64} \\
\SAGE    & \meanstd{34.65}{1.47} & \meanstd{48.75}{1.85} & \second{\meanstd{20.54}{1.40}} \\
\GAE     & \first{\meanstd{37.95}{1.24}} & \first{\meanstd{49.65}{1.48}} & \meanstd{10.50}{0.46} \\
\BUDDY   & \meanstd{30.40}{1.18} & \second{\meanstd{48.35}{1.18}} & \meanstd{16.78}{0.53} \\
\NEOGNN  & \meanstd{31.27}{0.72} & \meanstd{41.74}{1.18} & \meanstd{17.88}{0.71} \\
\NCN     & \meanstd{35.14}{1.04} & \third{\meanstd{53.41}{1.46}} & \meanstd{13.22}{0.56} \\
\NCNC    & \third{\meanstd{36.70}{1.57}} & \meanstd{53.72}{0.97} & \third{\meanstd{18.81}{1.16}} \\
\SEAL    & \meanstd{24.27}{6.74} & \meanstd{27.37}{3.20} & \meanstd{12.47}{1.23} \\
\bottomrule
\end{tabular}
\end{table}

\begin{table}[t]
\caption{Hits@20 with standard deviations on real-world link prediction datasets.}
\label{tab:hits20-part3}
\centering
\setlength{\tabcolsep}{4pt}
\footnotesize
\begin{tabular}{lcccc}
\toprule
\textbf{Models} & \textsc{ogbl-citation2} & \textsc{ogbl-ddi} & \textsc{ogbl-ppa} & \textsc{ogbl-collab} \\
$\hat{r}_G$ & 0.01 & 0.10 & 0.12 & 0.13 \\
\midrule
\GCN     & \second{\meanstd{51.72}{0.46}} & \meanstd{64.76}{1.45} & \meanstd{68.38}{0.73} & \meanstd{22.48}{0.81} \\
\GAT     & \first{\meanstd{53.13}{0.15}} & \second{\meanstd{66.83}{2.23}} & \meanstd{69.49}{0.43} & \meanstd{18.30}{1.42} \\
\SAGE    & \texttt{OOM} & \first{\meanstd{67.19}{1.18}} & \texttt{OOM} & \meanstd{21.26}{1.32} \\
\GAE     & \texttt{OOM} & \meanstd{17.81}{9.80} & \texttt{OOM} & \texttt{OOM} \\
\BUDDY   & \meanstd{47.81}{0.37} & \meanstd{58.71}{1.63} & \meanstd{71.50}{0.68} & \second{\meanstd{23.35}{0.73}} \\
\NEOGNN  & \meanstd{43.17}{0.53} & \meanstd{51.94}{10.33} & \meanstd{64.81}{2.26} & \meanstd{21.03}{3.39} \\
\NCN     & \third{\meanstd{53.76}{0.20}} & \third{\meanstd{65.82}{2.66}} & \first{\meanstd{81.89}{0.31}} & \meanstd{20.84}{1.31} \\
\NCNC    & \meanstd{51.69}{1.48} & \texttt{>24h} & \second{\meanstd{82.24}{0.40}} & \meanstd{20.49}{3.97} \\
\SEAL    & \meanstd{48.62}{1.93} & \meanstd{49.74}{2.39} & \third{\meanstd{76.77}{0.94}} & \meanstd{21.57}{0.38} \\
\bottomrule
\end{tabular}
\end{table}

\end{document}